\documentclass[letterpaper]{article} 
\usepackage{aaai2027}  
\usepackage[hyphens]{url}  
\usepackage{graphicx} 
\usepackage{natbib}
\usepackage{caption}
\usepackage{amssymb}
\usepackage{amsmath}
\usepackage{multirow}
\usepackage{algorithm}
\usepackage{algorithmic}

\usepackage{newfloat}
\usepackage{listings}
\DeclareCaptionStyle{ruled}{labelfont=normalfont,labelsep=colon,strut=off} 
\floatstyle{ruled}
\newfloat{listing}{tb}{lst}{}
\floatname{listing}{Listing}

\usepackage{booktabs}

\nocopyright
\title{Attention-Path Fragility as an Uncertainty Signal in Large Language Models}
\author{
    Minsoo Kim,
    Sungyoung Ji,
    Kisung Moon,
    Ilyong Yoon\corresponding
}
\affiliations{
    POSCO Holdings Future Technology Research Institute\\
    440, Teheran-ro, Gangnam-gu, Seoul, Republic of Korea\\
    \{kim1102, sy\_ji, moonkisung, ilyong.yoon\}@posco-inc.com
}

\begin{document}
\maketitle
\begin{abstract}
We propose that a model's uncertainty about a token is reflected not only in the breadth of its output distribution but also in whether a confident prediction is \emph{fragile} under perturbation of its attention pathways. We instantiate this as ASMI (Attention-Subnetwork Mutual Information), a training-free estimator that masks attention heads and measures the BALD mutual information among the resulting subnetworks, with a semantic-agreement kernel to discount surface-form disagreement. The signal is not a restatement of output confidence: on grounded QA an out-of-fold test shows it adds error-predictive information beyond single-pass confidence and entropy, concentrated in \emph{confident-but-fragile} predictions, where acting on it roughly halves the retained error of a confidence filter. The distinctness is regime-graded, so ASMI predicts its own domain of applicability, strong where answers are routed through provided context and bounded by design where they are recalled from parametric knowledge. Sem-ASMI reads the signal from a single greedy response, without the stochastic generations the strongest baselines require, and ties or beats Semantic Entropy on ten of the twelve grounded benchmark-backbone settings. Across the same twelve settings, the best ASMI variant, typically the adaptive one reusing the ten samples already drawn for the baselines, ties or leads the strongest baseline in eight, significantly in three under a paired test. On parametric QA all variants revert to or below the zero-cost MSP baseline, exactly as predicted, and the estimates are near-deterministic across reruns. A head-level analysis shows that what tracks this boundary is not the presence of head-level fragility but whether that fragility couples to errors.
\end{abstract}
\section{Introduction}
\label{sec:intro}
Reliable uncertainty estimation (UE) is foundational for deploying large language models (LLMs) in selective generation, hallucination detection, and human-in-the-loop decision making~\citep{kadavath2022language,manakul2023selfcheckgpt,kalai2025language}, where over-confident incorrect outputs carry real costs~\citep{guo2017calibration,jiang2021how}. The stakes are highest in grounded generation, where a hallucination is a failure to route the answer through the provided context~\citep{huang2025survey}. Estimators span information-theoretic quantities on a single deterministic forward pass~\citep{malinin2020uncertainty}, agreement statistics across stochastically decoded samples~\citep{kuhn2023semantic,farquhar2024detecting,duan2024shifting,nikitin2024kernel}, signals read from internal states and from attention~\citep{azaria2023internal,vazhentsev2025rauq}, and structural perturbations of the model itself~\citep{gal2016dropout,zhang2025tokur}.
\begin{figure}[t!]
\centering
\includegraphics[width=1.0\linewidth]{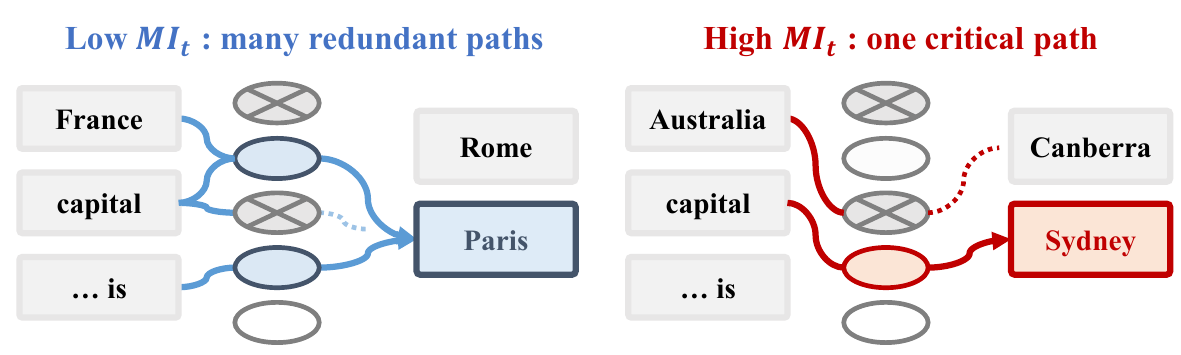}
\caption{ASMI at a glance. Masking heads leaves a confident token (left) unchanged but destabilizes a fragile token (right) that hinges on one critical path. ASMI scores this fragility: random attention-head masking at a target layer yields multiple predictive distributions whose disagreement, measured by mutual information, is a token-level uncertainty proxy.}
\label{fig:overview}
\end{figure}
We propose a complementary signal: a model's uncertainty about a token is reflected not only in the breadth of its output distribution, but also in whether the prediction is \emph{fragile}, that is, whether it hinges on a specific attention pathway among many redundant alternatives. The intuition rests on a well-documented property of multi-head attention: many heads can be pruned with minimal loss~\citep{michel2019sixteen,voita2019analyzing,he2024matters}, yet a high-level behavior can hinge on a single head~\citep{zhou2025role}, and a small subset is uncertainty-aware~\citep{vazhentsev2025rauq}. Masking leaves a genuinely supported prediction intact, since redundant pathways carry the same information, and destabilizes a fragile one (Fig.~\ref{fig:overview}). Crucially, this fragility is invisible to a single forward pass: a token can be assigned high confidence yet collapse under masking, and these confident-but-fragile tokens carry error information that output-confidence measures miss.
We instantiate this as ASMI (Attention-Subnetwork Mutual Information): for each Monte Carlo sample we mask attention heads at a target layer and quantify disagreement among the resulting predictive distributions via the BALD mutual-information decomposition~\citep{houlsby2011bald, smith2018uncertainty}, with a semantic-agreement kernel to discount surface-form disagreement. The estimator is training-free and needs only the choice of target layer.
Existing estimators can be characterized along two axes: their \emph{variation source} (stochastic output sampling versus structural model perturbation) and their \emph{measurement target} (output spread versus computational path dependence). ASMI occupies the structural-perturbation, path-dependence corner, and its position yields a testable hypothesis: the signal should be informative where correctness depends on attending to specific context tokens, as in retrieval-grounded QA, and end where uncertainty originates elsewhere, as in parametric knowledge recall~\citep{longpre2021entity,mallen2023trust,xu2024knowledge}. We test this behaviorally, with grounded against parametric QA as a designed contrast, and internally, asking whether the signal adds error information beyond output confidence and how its head-level signature tracks the boundary. The result is an estimator that states in advance where it should work and where it should fail, with both directions confirmed. Our contributions are:
\begin{itemize}
    \item We propose \textbf{ASMI}, a training-free token-level uncertainty estimator that probes attention-path fragility by random head masking with a semantic-agreement kernel, and reads it from a single greedy response. In this sampling-free form it is competitive with the strongest sample-diversity baselines, which each draw ten stochastic generations, and its estimates are near-deterministic across reruns.
    \item We show the signal is \textbf{distinct from output confidence, not a proxy for it}: an out-of-fold test confirms it adds error-predictive information beyond single-pass confidence and entropy, concentrated in \emph{confident-but-fragile} predictions.
    \item The distinctness is \textbf{regime-graded, so ASMI predicts its own domain of applicability}: it leads on context-routed QA and reverts to or below the free MSP baseline on parametric recall, exactly where the design predicts. A head-level analysis shows that what tracks this boundary is whether fragility couples to errors, present on grounded QA and absent under parametric recall, and a near-zero masking response identifies the one backbone where the signal carries no information, a candidate label-free screen whose validation here is retrospective.
\end{itemize}
\begin{figure*}[ht]
    \centering
    \includegraphics[width=0.9\linewidth]{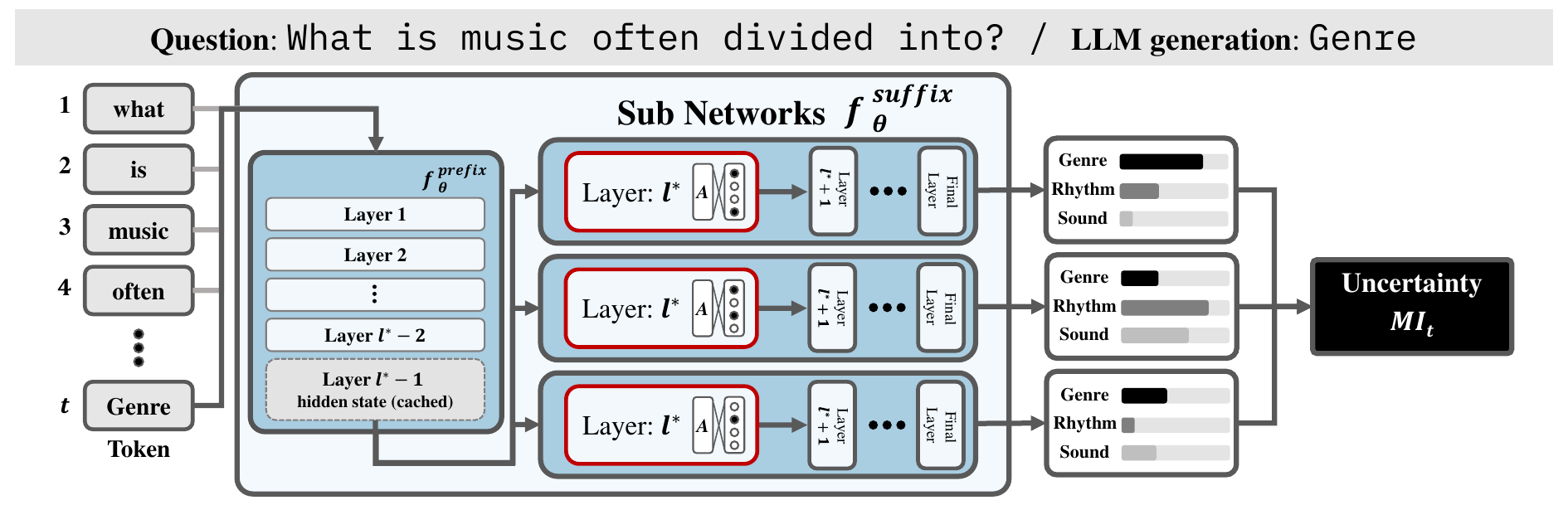}
    \caption{Pipeline of ASMI. The prefix below the target layer $\ell^{\star}$ is computed once, and $S$ independently masked copies of the suffix score the same greedy output (three shown). Filled and open circles denote kept and masked heads. Disagreement among the resulting token distributions yields $\mathrm{MI}_t$ (Eq.~\ref{eq:mi}). The semantic kernel of Sem-ASMI and the adaptive gate of Adapt-ASMI are omitted for clarity.}
    \label{fig:pipeline}
\end{figure*}
\section{Related Work}
\label{sec:related}
\paragraph{Sample-diversity UE.}
Sample-diversity methods estimate uncertainty from the variability of stochastically generated outputs. SelfCheckGPT~\citep{manakul2023selfcheckgpt} uses pairwise consistency among sampled generations as a hallucination signal. Semantic Entropy~\citep{kuhn2023semantic,farquhar2024detecting} clusters semantically equivalent generations and takes entropy over the clusters, Kernel Language Entropy~\citep{nikitin2024kernel} relaxes this to a soft semantic kernel, SAR~\citep{duan2024shifting} weights samples by relevance to the predicted answer, and LUQ~\citep{zhang2024luq} extends the approach to long-form text.
\paragraph{Single-pass UE.}
\emph{Information-based methods} such as Maximum Sequence Probability, Perplexity, and Mean Token Entropy~\citep{malinin2020uncertainty} read uncertainty directly from the output distribution of a single deterministic forward pass. \emph{Probing methods} examine internal model states: SAPLMA~\citep{azaria2023internal}, INSIDE~\citep{chen2024inside}, and Semantic Entropy Probes~\citep{kossen2024semantic} extract signals from hidden representations, while P(True)~\citep{kadavath2022language} prompts the model for self-assessment. \emph{Attention-based methods} such as RAUQ~\citep{vazhentsev2025rauq} read uncertainty from attention scores at heads identified as uncertainty-aware by prior analysis.
\paragraph{Structural perturbation UE.}
MC Dropout~\citep{gal2016dropout} established test-time structural perturbation as an uncertainty signal, but its variational interpretation does not transfer directly to modern LLMs trained without dropout, and adaptations such as TokUR~\citep{zhang2025tokur} inject noise into the attention weight matrices and read a posterior over parameters. ASMI perturbs differently: we apply discrete Bernoulli masks to attention head outputs to probe path dependence among redundant heads~\citep{michel2019sixteen,voita2019analyzing}, surfacing the model's reliance on specific attention paths rather than posterior uncertainty over parameters.
\paragraph{Adaptive UE.}
Other adaptive UE approaches such as Cocoa~\citep{vashurin2025cocoa} and GENUINE~\citep{wang2025genuine} combine multiple uncertainty signals via calibration or graph-based fusion. Our adaptive variant, Adapt-ASMI, instead modulates a single signal by a directly observable input property, the diversity of stochastically decoded outputs.
\paragraph{A two-axis characterization.}
Placing the families above on those two axes leaves one corner empty. Sample-diversity methods measure output spread, information-based and probing methods read outputs or hidden states without perturbing, RAUQ and Lookback Lens~\citep{chuang2024lookback} observe attention passively, and MC Dropout perturbs the model but still measures output spread. ASMI is, to our knowledge, the first estimator in the structural-perturbation, path-dependence corner, and that position is predictive rather than merely descriptive. It should be informative when the prediction depends on routing context through specific attention paths, a computation carried by a sparse set of heads that causally transport in-context information~\citep{olsson2022context,wu2025retrieval,todd2024function} and can individually flip the answer between its in-context and memorized source~\citep{yu2023characterizing,jin2024cutting,ortu2024competition}. That advantage should end where uncertainty originates in parametric recall, whose locus is MLP-stored knowledge rather than the attention paths that extract it~\citep{geva2021transformer,dai2022knowledge,meng2022locating,geva2023dissecting}, a dissociation already exploited to detect RAG hallucination~\citep{sun2025redeep}. The hypothesis is falsifiable in both directions: it fails if the signal survives closed-book recall or if matched perturbations of non-attention components at the same layer reproduce it.
\section{Methodology}
\label{sec:method}
\begin{figure}[t!]
\centering
\includegraphics[width=0.52\linewidth]{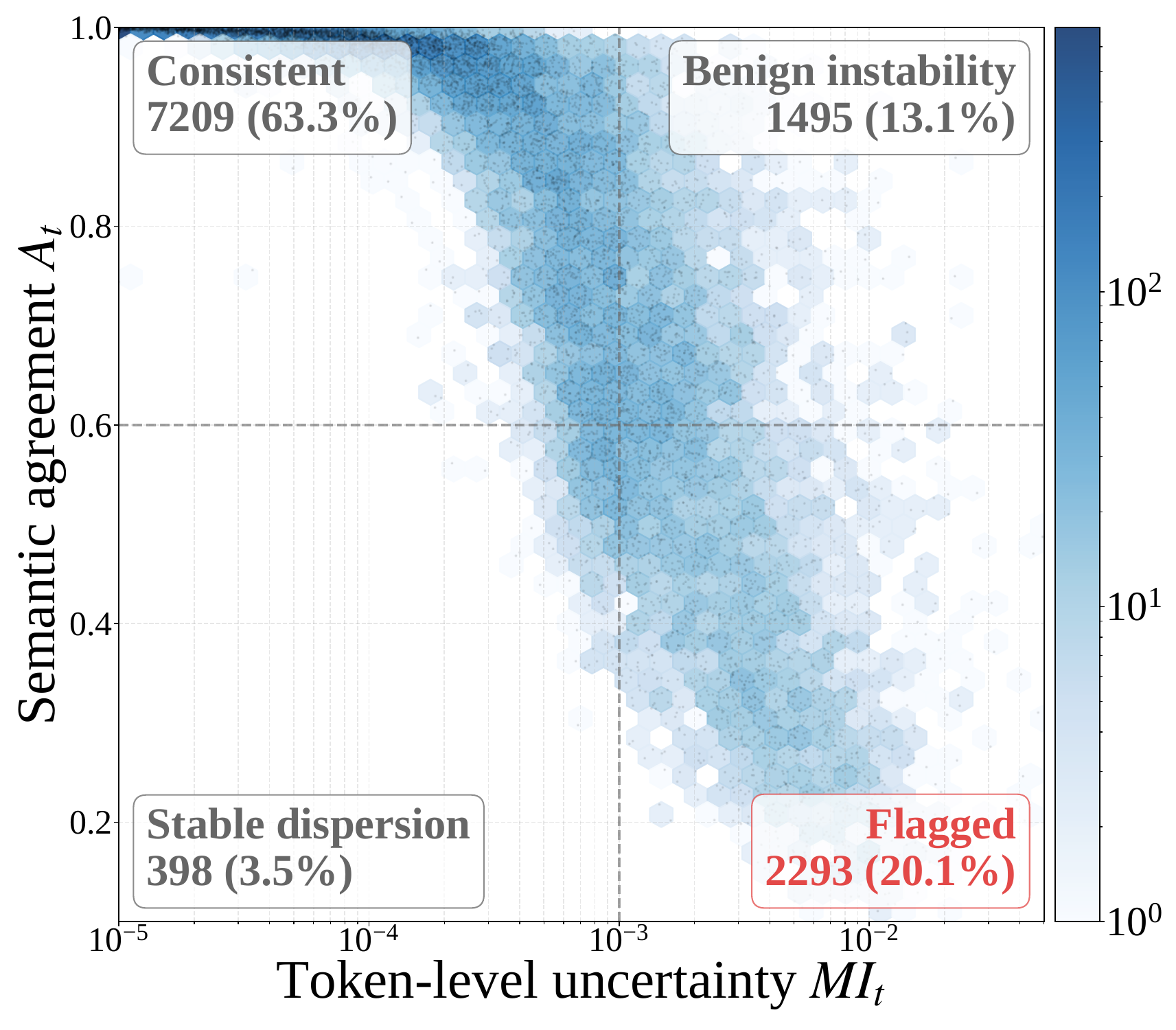}
\hfill
\includegraphics[width=0.47\linewidth]{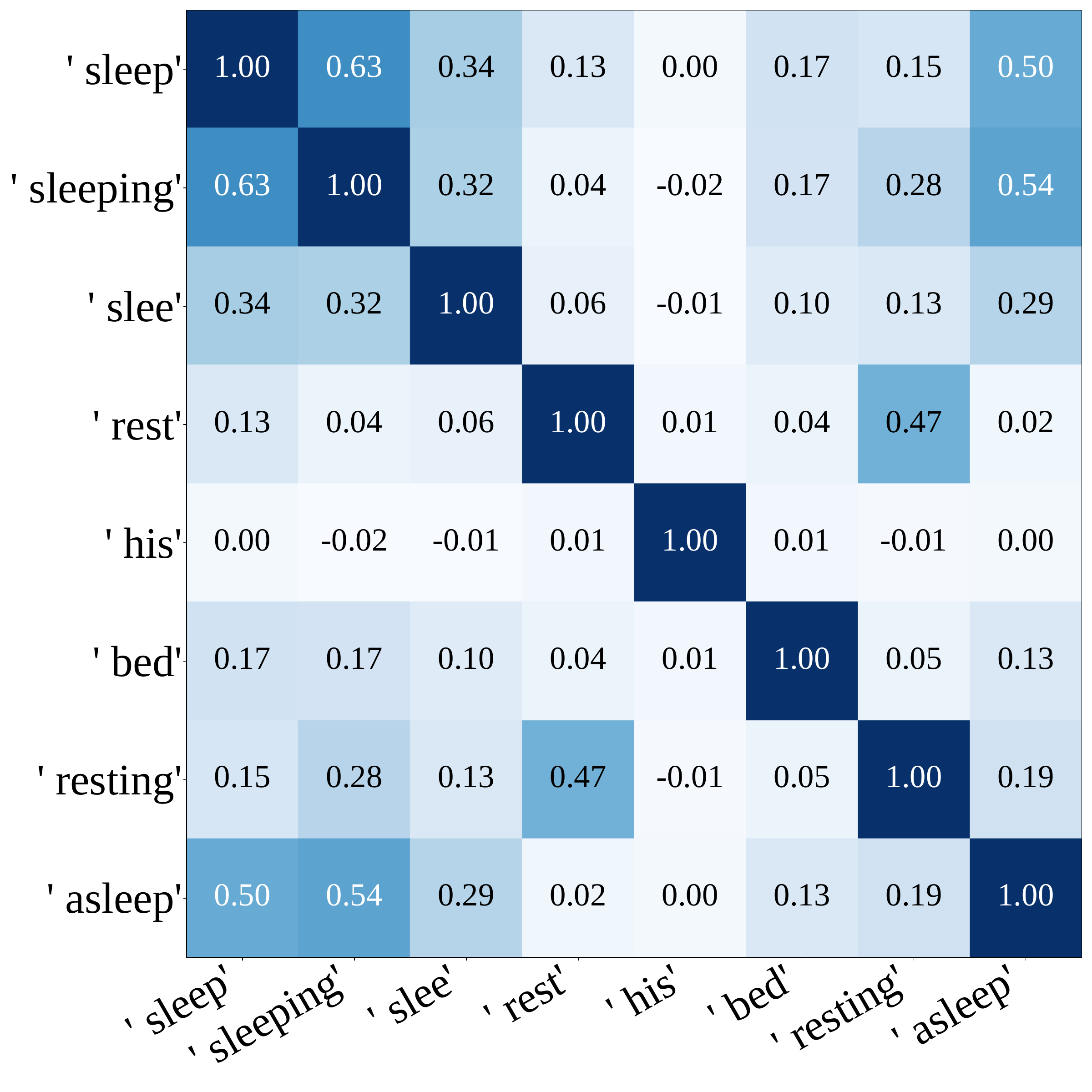}
\caption{Token-level mechanism on CoQA. \textbf{(a)} Joint distribution of $\mathrm{MI}_t$ and $A_t$ (Pearson $r = -0.57$). High $\mathrm{MI}_t$ with low agreement is flagged, while high $\mathrm{MI}_t$ with high agreement is benign instability that the kernel discounts. \textbf{(b)} Pairwise $W_{\mathrm{lm}}$ cosine similarity among the top-8 candidates at a representative benign-instability token, the surface-form cluster that drives $A_t$ up.}
\label{fig:mechanism}
\end{figure}

\paragraph{Attention subnetwork sampling.}
Let $f_\theta$ be a pretrained autoregressive language model with fixed parameters $\theta$. For an input context $(x, y_{<t})$, the model defines a next-token distribution $p_\theta(y_t \mid x, y_{<t})$. To estimate structural uncertainty from internal computation rather than decoding noise, we perturb the model at inference time by masking attention heads in a selected Transformer layer $\ell^\star$. For a multi-head attention module with $H$ heads,
\begin{equation}
\mathrm{MHA}(Q, K, V) = \mathrm{Concat}(\mathrm{head}_1, \ldots, \mathrm{head}_H)\, W_O,
\end{equation}
where
\begin{equation}
\mathrm{head}_h = \mathrm{softmax}\!\left( \frac{Q_h K_h^\top}{\sqrt{d_h}} \right) V_h.
\end{equation}
Because many heads can be pruned without significant loss~\citep{michel2019sixteen,voita2019analyzing,he2024matters}, we use random head dropout in reverse, as a probe of how strongly a prediction depends on specific attention paths. For each Monte Carlo sample $s$, we draw a binary mask $m^{(s)} \in \{0,1\}^H$ with
\begin{equation}
m^{(s)}_h \sim \mathrm{Bernoulli}(1 - p),
\end{equation}
and apply it to the head outputs before the output projection:
\begin{equation}
\widehat{\mathrm{MHA}}^{(s)}(Q, K, V) = \mathrm{Concat}\!\left( \widehat{\mathrm{head}}^{(s)}_1, \ldots, \widehat{\mathrm{head}}^{(s)}_H \right) W_O.
\end{equation}
No $1/(1-p)$ rescaling is applied, following the ablation convention of the head-pruning literature~\citep{michel2019sixteen,voita2019analyzing}, and the scale reduction is common to all samples, so it does not itself generate the inter-sample disagreement that $\mathrm{MI}_t$ measures. Because only the aggregated head outputs are masked, the attention routing remains fixed and only each head's contribution to the representation is perturbed.
We set $(p, S) = (0.15, 40)$, derived from a coverage lower bound
on $S$ and a fidelity upper bound on $p$
(Appendix~\ref{app:coverage-fidelity}).
\paragraph{Efficient Monte Carlo inference.}
Because the perturbation is applied only at layer $\ell^\star$, the
computation below that layer is shared across all $S$ subnetworks (Fig.~\ref{fig:pipeline}). Writing
the model as $f_\theta = f^{\mathrm{suf}}_\theta \circ
f^{\mathrm{pre}}_\theta$, where $f^{\mathrm{pre}}_\theta$ contains layers
$1, \ldots, \ell^\star - 1$ and $f^{\mathrm{suf}}_\theta$ contains layers
$\ell^\star, \ldots, L$, we compute the prefix activations
$h^{\mathrm{cache}} = f^{\mathrm{pre}}_\theta(x, y)$ once and evaluate only
the suffix under each mask $m^{(s)}$. This reduces the Monte Carlo cost from
$S(C_{\mathrm{pre}} + C_{\mathrm{suf}})$ to $C_{\mathrm{pre}} +
S\,C_{\mathrm{suf}}$ FLOPs, where $C_{\mathrm{pre}}$ and
$C_{\mathrm{suf}}$ denote prefix and suffix costs. Since $y$ is fixed prior
to scoring, each masked pass is a single teacher-forcing evaluation, and the
ensemble is defined implicitly by the binary masks: no parameter copies are
materialized, and the only persistent overhead is the shared cache
$h^{\mathrm{cache}}$ (measured overhead in
Appendix~\ref{app:impl-details}).
\paragraph{Mutual-information uncertainty.}
Let $p^{(s)}_t(\cdot)$ denote the predictive distribution at position $t$ under the $s$-th sampled attention subnetwork, and let $S$ be the number of Monte Carlo samples. We quantify token-level disagreement by the mutual information between the sampled subnetwork and the next-token prediction~\citep{houlsby2011bald}:
\begin{equation}
\mathrm{MI}_t = H(\bar{p}_t) - \frac{1}{S} \sum_{s=1}^{S} H\!\left( p^{(s)}_t \right), \quad \bar{p}_t = \frac{1}{S} \sum_{s=1}^{S} p^{(s)}_t,
\label{eq:mi}
\end{equation}
where $H(p) = -\sum_i p_i \log p_i$. This quantity becomes large when different attention subnetworks make confident but inconsistent predictions, and therefore captures structural disagreement induced by internal attention perturbations rather than decoding stochasticity.
Exact $\mathrm{MI}_t$ requires entropies over the full vocabulary ($|V| > 150K$), prohibitive across $S$ samples and all generated positions, so we approximate each $p^{(s)}_t$ by its top-$K$ candidates plus a single tail bucket holding the residual mass. On retrieval-grounded benchmarks top-$K = 64$ retains over 95\% of probability mass even at the most uncertain high-MI tokens (Appendix~\ref{app:topk-coverage}).
For sample $s$, let $\mathcal{I}^{(s)}_{t,K}$ be the top-$K$ token indices at position $t$, where each retained probability is computed using the full-vocabulary normalization constant:
\begin{equation}
p^{(s)}_t(i) = \frac{\exp(\ell^{(s)}_{t,i})}{\sum_{j=1}^{V} \exp(\ell^{(s)}_{t,j})}.
\end{equation}
We then form the union support
\begin{equation}
\mathcal{U}_t = \bigcup_{s=1}^{S} \mathcal{I}^{(s)}_{t,K},
\end{equation}
and construct a reduced distribution $\tilde{p}^{(s)}_t$ over $\mathcal{U}_t \cup \{\mathrm{tail}\}$:
\begin{equation}
\tilde{p}^{(s)}_t(i) =
\begin{cases}
p^{(s)}_t(i), & i \in \mathcal{U}_t, \\
1 - \sum_{j \in \mathcal{U}_t} p^{(s)}_t(j), & i = \mathrm{tail}.
\end{cases}
\end{equation}
We use these reduced distributions to approximate $\mathrm{MI}_t$. Eq.~\ref{eq:mi} admits the equivalent form
\begin{equation}
\mathrm{MI}_t \;=\; \frac{1}{S} \sum_{s=1}^{S} \mathrm{KL}\!\left( p^{(s)}_t \,\Big\|\, \bar{p}_t \right),
\label{eq:mi-kl}
\end{equation}
i.e., the average divergence of each subnetwork's prediction from the
ensemble mean. This form makes $\mathrm{MI}_t$ decomposable over the
masked heads.
\begin{algorithm}[t]
\caption{Sem-ASMI Uncertainty Estimation}
\label{alg:swasmi}
\begin{algorithmic}[1]
\REQUIRE Input $x$; model $f_\theta$; layer $\ell^\star$; samples $S$; mask rate $p$; truncation $K$
\ENSURE Sequence uncertainty $U(x, y)$
\STATE $y \leftarrow \mathrm{GreedyDecode}(f_\theta, x)$, \quad $T \leftarrow |y|$
\STATE $h^{\mathrm{cache}} \leftarrow f^{\mathrm{pre}}_\theta(x, y)$
\FOR{$s = 1$ \TO $S$}
    \STATE Sample mask $m^{(s)} \in \{0,1\}^H$, $m^{(s)}_h \sim \mathrm{Bernoulli}(1-p)$
    \STATE $\{p^{(s)}_t\}_{t=1}^{T} \leftarrow f^{\mathrm{suf}}_\theta(h^{\mathrm{cache}}; m^{(s)})$ at positions of $y$
\ENDFOR
\STATE Form top-$K$ union support and tail-bucket distributions $\{\tilde{p}^{(s)}_t\}$
\STATE Compute $\{\mathrm{MI}_t\}, \{A_t\}$ from $\{\tilde{p}^{(s)}_t\}$ via Eq.~\ref{eq:mi}, \ref{eq:agreement}
\STATE \textbf{return} $U \leftarrow \frac{1}{T} \sum_{t=1}^{T} \mathrm{MI}_t\,(1 - A_t)$
\end{algorithmic}
\end{algorithm}
\paragraph{Semantic agreement weighting.}
Different subnetworks may assign high probability to semantically similar alternatives (e.g., ``sleep'' and ``sleeping''), disagreement that is not uncertainty. To discount it, we compute a semantic agreement score on the top-$K$ candidate distributions (Fig.~\ref{fig:mechanism}). For MC samples $m$ and $n$, let $p^{(m)}_{t},\,p^{(n)}_{t}\in\mathbb{R}^{K}$ be their top-$K$ probability vectors,
renormalized to sum to one over the retained support, and let $G^{(m,n)}_t \in \mathbb{R}^{K \times K}$ be the pairwise token similarity matrix with entries equal to $1$ for identical tokens and $\min(1, \max(0, \cos(e_i, e_j)))$ otherwise, computed over L2-normalized rows of the output projection matrix $W_{\mathrm{lm}}$, so that $A_t \in [0, 1]$ by construction. We define pairwise agreement as
\begin{equation}
A^{(m,n)}_t = \left( p^{(m)}_t \right)^\top G^{(m,n)}_t\, p^{(n)}_t,
\label{eq:pair-agreement}
\end{equation}
and average it across ordered sample pairs:
\begin{equation}
A_t = \frac{1}{S(S-1)} \sum_{m \neq n} A^{(m,n)}_t.
\label{eq:agreement}
\end{equation}
The semantically-weighted token-level score is
\begin{equation}
u^{\mathrm{sem}}_t = \mathrm{MI}_t\, (1 - A_t),
\label{eq:sem-mi}
\end{equation}
while the unweighted base variant, which we denote ASMI and use as an ablation, retains the raw $u^{\mathrm{nosem}}_t = \mathrm{MI}_t$. The sequence-level uncertainty is the mean over generated token positions:
\begin{equation}
U(x, y) = \frac{1}{T} \sum_{t=1}^{T} u^{\mathrm{sem}}_t.
\label{eq:seq-unc}
\end{equation}
Equations~\ref{eq:sem-mi} and~\ref{eq:seq-unc} define Sem-ASMI.
\paragraph{Adaptive semantic weighting.}
Semantic weighting helps when valid answers share a canonical form, but risks discarding genuine uncertainty when many surface forms are valid. Adapt-ASMI therefore gates it per input by the diversity of $N$ sampled responses, each embedded by the model's own hidden state at layer $\lfloor L/2 \rfloor$ at its final token. With sample embeddings $e_1, \ldots, e_N$,
\begin{equation}
\mathrm{div}(x) = 1 - \tbinom{N}{2}^{-1} \textstyle\sum_{i<j} \cos(e_i, e_j),
\label{eq:diversity}
\end{equation}
and the gate is $\alpha(x)=\sigma\big(\beta(\tau-\mathrm{div}(x))\big)$ with
logistic $\sigma$, so that $\alpha\to 1$ when the sampled responses are near-paraphrases
($\mathrm{div}(x)<\tau$) and $\alpha\to 0$ when they genuinely diverge. The token-level score becomes
\begin{equation}
u^{\mathrm{adapt}}_t = \mathrm{MI}_t \bigl(1 - \alpha(x)\, A_t\bigr),
\label{eq:adapt-mi}
\end{equation}
recovering Eq.~\ref{eq:sem-mi} as $\alpha \to 1$ (low diversity) and the raw $\mathrm{MI}_t$ as $\alpha \to 0$, with $(\tau, \beta) = (0.3, 10)$ fixed across all tasks and backbones.
\begin{table*}[t]
\centering
\setlength{\tabcolsep}{2pt}
\resizebox{\textwidth}{!}{%
\begin{tabular}{l|cccc|cccc|cccc|cccc} \toprule
\multirow{2}{*}{UE Method} & \multicolumn{4}{c|}{Qwen3-4B} & \multicolumn{4}{c|}{Qwen3-8B} & \multicolumn{4}{c|}{Llama-2-7B} & \multicolumn{4}{c}{Mistral-7B} \\
\cline{2-17}
 & CoQA & SQuAD & BabiQA & TriviaQA & CoQA & SQuAD & BabiQA & TriviaQA & CoQA & SQuAD & BabiQA & TriviaQA & CoQA & SQuAD & BabiQA & TriviaQA \\
\hline\hline
\multicolumn{17}{l}{\textit{Baselines}} \\
Maximum Sequence Probability & 0.45 & 0.32 & 0.48 & \underline{0.72} & 0.53 & 0.35 & 0.76 & \underline{0.70} & 0.53 & 0.18 & \textbf{0.58} & 0.77 & 0.52 & 0.65 & 0.59 & 0.78 \\
Mean Token Entropy & 0.40 & 0.27 & 0.56 & 0.70 & 0.50 & 0.41 & 0.75 & 0.69 & 0.46 & 0.68 & 0.30 & 0.75 & 0.44 & 0.78 & 0.54 & 0.78 \\
Semantic Entropy & 0.46 & 0.28 & 0.54 & 0.72 & 0.53 & \underline{0.45} & 0.74 & 0.70 & \underline{0.55} & 0.47 & 0.50 & \underline{0.77} & \underline{0.54} & 0.80 & 0.58 & \underline{0.78} \\
SAR & 0.46 & 0.27 & \underline{0.63} & 0.71 & 0.47 & 0.40 & 0.71 & 0.68 & 0.53 & 0.51 & 0.49 & \textbf{0.78} & 0.52 & 0.78 & 0.56 & \textbf{0.80} \\
P(True) & 0.03 & 0.07 & 0.49 & 0.30 & 0.07 & 0.17 & 0.22 & 0.39 & -0.04 & 0.47 & 0.21 & 0.06 & -0.05 & 0.78 & 0.20 & -0.11 \\
RAUQ & 0.37 & \underline{0.32} & 0.32 & 0.70 & 0.48 & 0.45 & 0.74 & 0.66 & 0.47 & 0.46 & \underline{0.57} & 0.75 & 0.46 & 0.82 & 0.48 & 0.70 \\
RAUQ (entropy) & 0.47 & \textbf{0.36} & 0.42 & \textbf{0.72} & 0.53 & \textbf{0.47} & 0.74 & \textbf{0.70} & 0.50 & \textbf{0.75} & 0.29 & 0.77 & 0.51 & \underline{0.83} & 0.46 & 0.77 \\
\hline\hline
\multicolumn{17}{l}{\textit{Attention-based, active perturbation (ours)}} \\
ASMI (60\%) & 0.46 & 0.29 & 0.60 & 0.59 & 0.51 & 0.34 & \textbf{0.80} & 0.62 & 0.49 & 0.62 & 0.11 & 0.66 & 0.48 & 0.77 & 0.52 & 0.74 \\
Sem-ASMI (60\%) & \underline{0.52} & 0.29 & \textbf{0.66} & 0.66 & \underline{0.55} & 0.40 & 0.78 & 0.66 & 0.54 & 0.69 & 0.27 & 0.71 & 0.53 & 0.83 & \underline{0.61} & 0.78 \\
Adapt-ASMI (60\%) & \textbf{0.53} & 0.31 & \textbf{0.66} & 0.66 & \textbf{0.56} & 0.40 & \underline{0.79} & 0.67 & \textbf{0.55} & \underline{0.71} & 0.29 & 0.72 & \textbf{0.54} & \textbf{0.84} & \textbf{0.61} & 0.78 \\
\bottomrule
\end{tabular}
}
\caption{Representative UE methods on four backbones (PRR, higher is better, ASMI variants at depth $d = 60\%$). TriviaQA is the closed-book parametric control, and the other three benchmarks are context-routed. Bold and underline mark best and second best among the methods shown. Standard errors are $\pm 0.01$ to $\pm 0.03$ ($\pm 0.03$ to $\pm 0.07$ on BabiQA), with per-cell values and all 17 baselines in Appendix~\ref{app:full-benchmark}.}
\vspace{2mm}
\label{tab:bench_main}
\end{table*}
\section{Experiments}
\begin{table}[ht!]
\centering
\setlength{\tabcolsep}{5pt}
\resizebox{0.75\columnwidth}{!}{%
\begin{tabular}{lrccc}
\toprule
Dataset & $n$ & Input & Gen & AlignScore \\
\midrule
CoQA     & $\approx8\mathrm{K}$ & $490$ & $4$  & $0.782$ \\
SQuAD    & $5\mathrm{K}$ & $183$ & $18$ & $0.869$ \\
BabiQA   & $1\mathrm{K}$ & $175$ & $3$  & $0.709$ \\
\midrule
TriviaQA & $5\mathrm{K}$ & $176$ & $4$  & $0.488$ \\
\bottomrule
\end{tabular}%
}
\caption{Dataset characteristics on Qwen3-4B-base. Median input and generation lengths in tokens, and mean AlignScore of the greedy answers. The three grounded datasets sit above the closed-book parametric control.}
\label{tab:datasets}
\end{table}
\paragraph{Datasets and quality metric.}
Our benchmark is a two-condition contrast dictated by the method's design. The context-routed condition uses CoQA~\citep{reddy2019coqa}, SQuAD~\citep{rajpurkar2016squad}, and BabiQA~\citep{weston2015babi} (Table~\ref{tab:datasets}). As a negative control, the parametric condition uses \emph{closed-book} TriviaQA~\citep{joshi2017triviaqa} with no evidence documents, so answers must come from parametric memory~\citep{mallen2023trust}. All datasets follow the LM-Polygraph~\citep{vashurin2025benchmarking} pipeline with its default prompting and standard splits, scored by AlignScore~\citep{zha2023alignscore}, an entailment-based metric that tracks faithfulness better than surface-overlap scores~\citep{maynez2020faithfulness}. To bound compute, SQuAD and TriviaQA use a fixed random $5$K subsample, identical across methods and backbones.
\paragraph{Baselines.}
We compare against 17 baseline UE methods spanning four families. \emph{Information-based}: MSP, Perplexity, Mean Token Entropy~\citep{malinin2020uncertainty}, PMI~\citep{takayama2019pmi}, Conditional PMI~\citep{vanderpoel2022cpmi}, SelfCertainty~\citep{kang2025selfcertainty}. \emph{Sample-diversity}: Monte Carlo and MC-normalized sequence entropy~\citep{malinin2020uncertainty}, Semantic Entropy~\citep{kuhn2023semantic,farquhar2024detecting}, SAR, SentenceSAR~\citep{duan2024shifting}, LUQ~\citep{zhang2024luq}, Kernel Language Entropy~\citep{nikitin2024kernel}. \emph{Probing}: P(True) and its sampling variant~\citep{kadavath2022language}. \emph{Attention-based}: RAUQ and its entropy variant~\citep{vazhentsev2025rauq}. All run within LM-Polygraph with default hyperparameters. The main table shows a representative subset, and the full comparison with standard errors is in Appendix~\ref{app:full-benchmark}.
\paragraph{Implementation details.}
Our primary backbone is Qwen3-4B-base~\citep{yang2025qwen3}, with Qwen3-8B-base, Llama-2-7B~\citep{touvron2023llama2}, and Mistral-7B~\citep{jiang2023mistral7b} for generality, all with $H = 32$ query heads. We evaluate base models deliberately, since instruction tuning reshapes attention-head specialization~\citep{wu2024language} and would confound the head-level analysis. We fix a single operating point $(p, S) = (0.15, 40)$ with top-$K$ truncation $K = 64$ across all tasks and backbones (Appendix~\ref{app:coverage-fidelity}). The estimator's only hyperparameter is the masked layer, a relative depth $d$ that transfers across architectures. Because the context-routing computation ASMI probes concentrates in the middle to later-middle layers~\citep{geva2023dissecting,wu2025retrieval}, we sweep $d \in \{60, 70, 80, 90\}\%$ identically everywhere and take $d = 60\%$ as representative by aggregate PRR, which needs no per-task tuning (Appendix~\ref{app:full-sweep}). Responses are greedy-decoded and scored under each mask. Adapt-ASMI's gate reuses the $N = 10$ sampled responses the protocol already generates for the sampling baselines, whereas ASMI and Sem-ASMI need only the greedy response. All methods are evaluated by the Prediction Rejection Ratio (PRR)~\citep{malinin2020uncertainty} within LM-Polygraph~\citep{fadeeva2023lm}, the normalized area between the uncertainty-based and random rejection curves, taking the value $1$ for oracle rejection and $0$ for an uninformative score (Appendix~\ref{app:impl-details}).

\begin{table}[t!]
\centering
\setlength{\tabcolsep}{6pt}
\resizebox{0.95\columnwidth}{!}{%
\begin{tabular}{lcc}
\toprule
Wall-clock s/example (median) & CoQA & SQuAD \\
\midrule
MSP & $4.14$ & $0.75$ \\
Semantic Entropy (10 samples + NLI) & $8.73$ & $2.16$ \\
Sem-ASMI (greedy + 40 masked passes) & $6.35$ & $2.48$ \\
Adapt-ASMI (adds the 10 samples) & $10.77$ & $3.76$ \\
\midrule
Sem-ASMI, marginal given the samples & $2.21$ & $1.74$ \\
\bottomrule
\end{tabular}
}
\caption{End-to-end cost on Qwen3-4B-base (200 examples, one H100 MIG slice, mask-batched Sem-ASMI). Full breakdown in Appendix~\ref{app:cost}.}
\label{tab:cost}
\end{table}
\paragraph{Main results.}
Table~\ref{tab:bench_main} is best read together with measured cost (Table~\ref{tab:cost}). Sem-ASMI needs only the greedy response, and in that sampling-free form it is competitive on grounded QA with the strongest sample-diversity baselines, which each draw ten stochastic generations per input. Measured cost tracks input length and therefore favors the long-context inputs typical of context-routed QA: on CoQA the mask-batched implementation takes $27\%$ less wall-clock end to end than Semantic Entropy, on the shorter SQuAD inputs it takes $15\%$ more, and once the ten samples exist the marginal cost of the score itself is $2.2$ and $1.7$ seconds. Under a cluster-respecting paired bootstrap the best ASMI variant ties or leads the per-column best baseline in eight of the twelve grounded columns, significantly in three, and in every remaining lead the sampling-free Sem-ASMI ties or beats Semantic Entropy from the greedy response alone (Appendix~\ref{app:h2h}). The same test finds four significant deficits, confined to SQuAD, where three of the four backbones trail the per-column best baseline, and to the one mapped exception cell below. On parametric recall all variants revert to or below the free MSP baseline, and the paired test places the top variant significantly below the per-column best baseline in three of the four columns, so the same table draws the boundary it was built to test. The three ASMI rows isolate the ablation: the semantic kernel lifts Sem-ASMI over the unweighted ASMI by $0.04$ to $0.06$ PRR on CoQA, where surface forms vary most, and the adaptive gate matches or exceeds it elsewhere. A further ablation shows the gain is not the averaging: the ensemble-mean entropy from the same forty masked passes is rank-identical to single-pass entropy (Spearman $0.97$ to $0.995$), and the weighted score beats it significantly on both grounded analysis cells (Appendix~\ref{app:distinct}). On CoQA the semantically weighted variants, Sem- and Adapt-ASMI, lead both RAUQ and its entropy variant on all four backbones, and the gap between those two RAUQ variants shows that RAUQ's competitiveness comes from its entropy term rather than from attention, so the attention signal ASMI reads is not already captured by the strongest attention baseline. ASMI is also the more reproducible of the two families: redrawing only internal randomness moves Adapt-ASMI's PRR by $\pm 0.006$ against $\pm 0.014$ to $0.041$ for the sample-diversity competitors (Appendix~\ref{app:stability}). Absolute PRR is not comparable across columns, so all comparisons are within a column.
\begin{figure}[t!]
\centering
\includegraphics[width=0.50\linewidth]{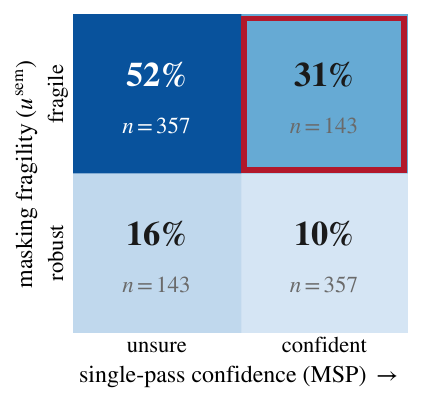}
\hfill
\includegraphics[width=0.49\linewidth]{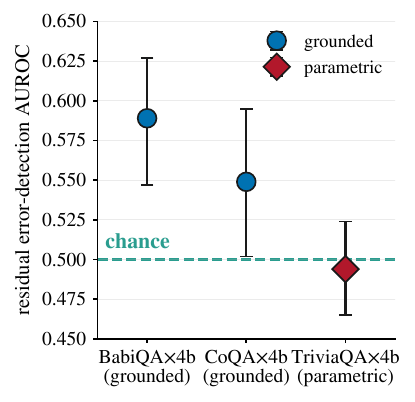}
\caption{The confident-but-fragile signal. \textbf{(a)} BabiQA error rate by MSP confidence and $u^{\mathrm{sem}}$ fragility: confident-but-fragile predictions fail $31\%$ against $10\%$ for confident-robust. \textbf{(b)} The effect is strong on grounded QA and fades to chance under parametric recall.}
\label{fig:confident-fragile}
\end{figure}
\paragraph{A distinct signal, not a proxy for output confidence.}
Aggregate PRR shows ASMI ranks errors as well as the strongest single-pass baselines, but not that it reads a different signal. Because token-level MI and single-pass entropy are strongly correlated (Spearman around $0.9$ on CoQA), we test distinctness at the source rather than the aggregate. On the greedy response we regress the semantically-weighted score $u^{\mathrm{sem}}$ on MSP and single-pass entropy out of fold and ask whether its residual still predicts errors. It does. The residual detects errors above chance on every grounded dataset, in five of the six backbone-dataset cells at AUROCs of $0.54$ to $0.60$, one of them at the estimator's own Monte Carlo resolution, so the information $u^{\mathrm{sem}}$ adds is not contained in output confidence or entropy (Table~\ref{tab:app-distinct}), consistent with theoretical and probing evidence that output confidence leaves error information unread~\citep{yadkori2024believe,yuksekgonul2024attention,orgad2024llms}. The effect is localized: splitting sequences at the medians of MSP and $u^{\mathrm{sem}}$, the confident-but-fragile cell is wrong three times as often as the confident-robust cell, on BabiQA $31\%$ against $10\%$, a gap MSP alone cannot see (Fig.~\ref{fig:confident-fragile}). The signal is regime-graded exactly as the design predicts: the raw gap between fragile and robust cells also appears on parametric TriviaQA, but there it is entirely absorbed by single-pass confidence and entropy with the residual at chance on both backbones ($0.49$ and $0.51$), whereas on grounded QA a residual beyond both remains (Table~\ref{tab:app-distinct}). A separate check on real grounded QA shows the semantic kernel improving the ranking by discounting surface-form variation rather than adding information orthogonal to the masking disagreement, so there the distinct signal is carried by the attention-masking disagreement itself (Appendix~\ref{app:distinct}).
\begin{figure}[t!]
\centering
\includegraphics[width=\linewidth]{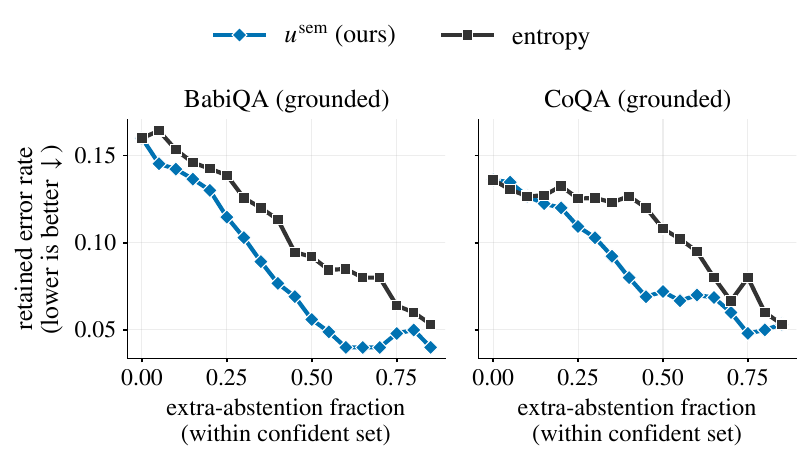}
\caption{Deployment payoff in the confident stratum. Retained error among MSP-confident predictions as an increasing fraction of them is abstained by ASMI fragility ($u^{\mathrm{sem}}$, blue) or by predictive entropy (black). On grounded QA the ASMI curve stays below entropy at every abstention level.}
\label{fig:payoff}
\end{figure}
\paragraph{Deployment payoff.}
In selective prediction~\citep{geifman2017selective,kamath2020selective}, the value of the signal concentrates on confident errors, the predictions a standard confidence filter keeps and acts on. Among the answers MSP rates as confident, abstaining the ASMI-fragile ones roughly halves the retained error on grounded QA, from $16.0\%$ to $5.6\%$ on BabiQA and from $13.6\%$ to $7.2\%$ on CoQA, against $9.2\%$ and $10.8\%$ for the same abstention budget spent on entropy, and the gap holds across the whole abstention range rather than at a single operating point (Table~\ref{tab:payoff}, Fig.~\ref{fig:payoff}). On parametric TriviaQA the order reverses, so the boundary holds in deployment too. As a standalone score ASMI also gives a lower area under the risk-coverage curve than MSP or entropy on the grounded benchmarks with the ordering reversing on TriviaQA as the boundary predicts (Appendix~\ref{app:payoff}). The gain is localized rather than global: added to a logistic selector over MSP and entropy fit across all coverage levels, ASMI leaves the overall risk-coverage curve unchanged, because its error information concentrates in the confident stratum and a single global fit averages that region together with the rest. A case-level look matches the mechanism. For confident BabiQA errors most masked subnetworks reproduce the confident wrong answer while a minority route to the true location, recovering garden where the greedy answer is bedroom, a disagreement the single pass and its entropy cannot see.
\begin{table}[t]
\centering
\setlength{\tabcolsep}{6pt}
\resizebox{1.0\columnwidth}{!}{%
\begin{tabular}{lccc}
\toprule
 & Confident error & Filter by ASMI & Filter by entropy \\
\midrule
BabiQA   & $0.160$ & $\mathbf{0.056}$ & $0.092$ \\
CoQA     & $0.136$ & $\mathbf{0.072}$ & $0.108$ \\
TriviaQA & $0.344$ & $0.147$ & $\mathbf{0.133}$ \\
\bottomrule
\end{tabular}
}
\caption{Retained error in the MSP-confident stratum after abstaining its top half by $u^{\mathrm{sem}}$ or by entropy (final coverage $0.25$, Qwen3-4B-base). On grounded QA ASMI roughly halves the confident error, while on parametric TriviaQA entropy leads instead.}
\label{tab:payoff}
\end{table}
\paragraph{Boundary: parametric-knowledge QA.}
TriviaQA is the designed negative control, and the boundary is sharp. On the same SQuAD examples with the passage removed, Sem-ASMI sits significantly below MSP ($-0.054$, 95\% CI $[-0.088, -0.021]$) while the open-book condition is a statistical tie, so the pattern reproduces within a single dataset when only the knowledge source moves (Appendix~\ref{app:boundary}). The practical instruction is symmetric: use this sampling-free signal as a faithfulness check where answers are routed through provided context, and hand off to output-distribution baselines for closed-book recall.
\paragraph{Operating envelope.}
Every remaining exception is mapped, not noise. All TriviaQA cells sit at or below MSP as the boundary predicts, and the one grounded exception has an identified cause. On BabiQA with Llama-2-7B head masking barely moves the output, so correct and incorrect answers carry the same near-zero MI and ASMI has no fragility to read while MSP still ranks the errors, an over-robustness that is a model property rather than an artifact (Table~\ref{tab:babiqa-mechanism}). The BabiQA depth instability is detailed in Appendix~\ref{app:full-sweep}. The cell announces itself before any correctness label exists: the same masking response that produces the score also reports when a model is too robust for it to carry information. We report this as a candidate label-free screen rather than a validated one, since its support is a single cell examined after the fact. 
\paragraph{Structural analysis.}
A head-level decomposition confirms that the fragility ASMI reads is head borne (Appendix~\ref{app:mechanism-details}). At the operating layer a causal probe makes the dependence direct: ablating a token's few most critical heads flips the prediction for high-MI tokens while leaving low-MI tokens almost unchanged. The concentration of this dependence couples positively to uncertainty on the grounded benchmarks and negatively on parametric recall, and mutual information tracks the head structure more than single-pass entropy does.
\section{Limitations \& Future Work}
The BALD form of Eq.~\ref{eq:mi} admits an epistemic reading of $\mathrm{MI}_t$ under the variational view of inference-time dropout~\citep{gal2016dropout}, but entropy-based decompositions of epistemic and aleatoric uncertainty face formal objections~\citep{wimmer2023quantifying}, so we use $\mathrm{MI}_t$ as a disagreement functional and claim no empirical separation of the two. Distinctness replicates across both backbones and all three grounded datasets, with two caveats: the CoQA residual on Qwen3-8B does not exclude chance, and the SQuAD effect on Qwen3-4B lies at the estimator's own Monte Carlo resolution (Appendix~\ref{app:distinct}). On the parametric control the top-$K$ truncation retains $77.3\%$ of the high-MI mass against over $95\%$ on grounded data, but re-scoring TriviaQA at $K = 256$ raises mean token coverage to $0.97$ and leaves Sem-ASMI significantly below MSP, so the parametric-side deficit is not a truncation artifact (Appendix~\ref{app:boundary}). Extending the two-axis characterization to reasoning, open-ended generation, instruction-tuned models, and further parametric tasks is future work, for which our structural analysis provides the measurement tool.
\section{Conclusion}
We introduced attention-path fragility as an uncertainty signal, instantiated as ASMI, a training-free estimator that probes attention-pathway redundancy by random head masking and reads it from a single greedy response. Two correlated signals can look identical in aggregate, so we tested distinctness at the source: the semantically-weighted score adds error-predictive information beyond confidence and entropy, concentrated in confident-but-fragile predictions. Its value is local rather than global: a global selector cannot exploit it, yet in the confident stratum, where a deployed system acts, it roughly halves the error. The distinctness is regime-graded and confirmed by a designed contrast: from that single greedy pass ASMI is competitive with the strongest sampling-based baselines on context-routed QA, and reverts to or below the free MSP baseline on parametric recall, exactly as the hypothesis requires. A head-level analysis traces this boundary to whether fragility couples to errors, and a causal ablation confirms the fragility is carried by identifiable critical heads.
The lesson outlasts the estimator: for a signal tied to a specific computation, where it applies and where its value lands follow from the locus of a model's errors, and we offer ASMI as a case for treating a UE signal's domain of applicability as a predictable property rather than a discovered one.

\bibliography{aaai2027}
\appendix
\clearpage
\setcounter{secnumdepth}{2}
\section{Implementation Details}
\label{app:impl-details}

\paragraph{Compute and memory overhead.}
With prefix caching, the scoring cost of ASMI is $C_{\mathrm{pre}} + S\,C_{\mathrm{suf}}$ rather than $S(C_{\mathrm{pre}} + C_{\mathrm{suf}})$, where the prefix below the target layer is computed once and the suffix is recomputed under each of the $S$ masks. With $S = 40$ the resulting overhead depends only on the target depth. On the 36-layer Qwen backbones the masked layer sits at $\ell^\star = -14$, so a 22-layer prefix is computed once and a 14-layer suffix is scored 40 times, giving $(22 + 40 \times 14)/36 \approx 16\times$ the FLOPs of a single unmasked scoring pass. On the 32-layer Llama-2-7B and Mistral-7B backbones the masked layer sits at $\ell^\star = -13$, giving $(19 + 40 \times 13)/32 \approx 17\times$. Memory overhead is limited to the shared prefix cache $h^{\mathrm{cache}}$, since each subnetwork is defined implicitly by its binary mask $m^{(s)}$ and no parameter copies are materialized.

\paragraph{Evaluation protocol.}
The Prediction Rejection Ratio (PRR)~\citep{malinin2020uncertainty} is the
normalized area between the uncertainty-based rejection curve and the
random-rejection baseline, taking values in $[-1, 1]$, where $1$
corresponds to oracle rejection (rejecting examples in order of decreasing
error) and $0$ to uninformative uncertainty. We compute PRR with the
LM-Polygraph~\citep{fadeeva2023lm} implementation, using
AlignScore~\citep{zha2023alignscore} as the sequence-level quality metric
throughout. PRR uses the continuous AlignScore, and the binary analyses
(error rates, AUROC, and the confident-stratum filters) binarize
correctness at $0.5$. Standard errors for all reported PRR values are estimated by
bootstrap resampling over test examples.

\paragraph{Baseline configuration.}
All baseline UE methods are run within LM-Polygraph with their default hyperparameters, except that all stochastic sampling uses a fixed temperature of $0.5$ with $10$ generations, following the sampling protocol of SAR~\citep{duan2024shifting} and identical across methods and backbones. Greedy-decoded responses used for scoring are shared across all methods on each benchmark, so quality-metric values are identical across estimators and PRR differences reflect only the uncertainty rankings.
\section{Top-$K$ Coverage Analysis}
\label{app:topk-coverage}
This appendix justifies the top-$K$ approximation used in computing the mutual information $\mathrm{MI}_t$ defined in the Methodology section. We chose $K = 64$ across all experiments based on the following analysis.
\paragraph{Setup.}
For each benchmark and each Monte Carlo sample $s \in \{1, \ldots, S\}$, we collect the per-token predictive distributions $p^{(s)}_t$ produced by Qwen3-4B-base under the sampled attention-head mask $m^{(s)}$ at the target layer $\ell^\star$. At each token position we compute the cumulative probability mass covered by the top-$K$ candidates,
\begin{equation}
\mathrm{Coverage}^{(s)}_t(K) = \sum_{i \in \mathrm{top}\text{-}K(t, s)} p^{(s)}_t(i),
\end{equation}
where $\mathrm{top}\text{-}K(t, s)$ denotes the top-$K$ indices of $p^{(s)}_t$, and average over all generated tokens and all $S = 40$ samples.
\paragraph{MI stratification.}
We stratify tokens by their $\mathrm{MI}_t$ value into three 10\% bands: \emph{low-MI} (bottom decile), \emph{mid-MI} (the central band, $P_{45}$ to $P_{55}$), and \emph{high-MI} (top decile). The high-MI band is the stringent case, since these tokens have the most spread
subnetwork distributions.

\begin{table}[t]
\centering
\small
\setlength{\tabcolsep}{5pt}
\begin{tabular}{lrccccc}
\toprule
Benchmark & $n$ & $K{=}8$ & $16$ & $32$ & $64$ & $128$ \\
\midrule
CoQA     & $567$  & $0.851$ & $0.903$ & $0.935$ & $0.955$ & $0.969$ \\
SQuAD    & $1480$ & $0.955$ & $0.972$ & $0.982$ & $0.987$ & $0.991$ \\
BabiQA   & $216$  & $0.950$ & $0.966$ & $0.975$ & $0.981$ & $0.986$ \\
TriviaQA & $397$  & $0.556$ & $0.633$ & $0.705$ & $0.773$ & $0.833$ \\
\bottomrule
\end{tabular}
\caption{Cumulative top-$K$ coverage on the high-MI stratum, the top 10\% of tokens by $\mathrm{MI}_t$. The three grounded benchmarks are well covered at $K = 64$, whereas TriviaQA is not.}
\label{tab:topk-highmi}
\end{table}

\paragraph{Coverage results.}
Low-MI tokens are trivially covered, with $K = 1$ alone above 99\% of mass on every benchmark, confirming that confident predictions are sharply peaked. For mid-MI tokens, coverage at $K = 64$ exceeds 99\% on all four benchmarks, with TriviaQA the lowest. The high-MI band is the stringent case (Table~\ref{tab:topk-highmi}). The three grounded benchmarks stay above 95\% at $K = 64$, while TriviaQA reaches 77.3\% and remains at 88.5\% even at $K = 256$.

\paragraph{Approximation bias.}
For the three grounded benchmarks the excluded high-MI tail mass is at most about 4.5\% (CoQA, the tightest case), which bounds its entropy contribution by $-0.045 \log 0.045 \approx 0.14$ nats on either $H(\bar{p}_t)$ or $H(p^{(s)}_t)$, well below the cross-token variation in $\mathrm{MI}_t$ used as the uncertainty signal. On TriviaQA the excluded tail is larger, so the absolute bias is larger, but the relative ranking of tokens by $\mathrm{MI}_t$, which is what PRR depends on, is preserved across the choice of $K$ in our checks (Appendix~\ref{app:boundary}).

\paragraph{Relation to the mechanism.}
The coverage pattern follows the same grounded and parametric split as the benchmark results. On CoQA, SQuAD, and BabiQA the candidate set at uncertain tokens is constrained by the input context, so subnetwork outputs stay concentrated on a moderate number of candidates even at high MI. On TriviaQA, parametric retrieval of factoid entities exposes a long tail of plausible alternatives, and coverage rises slowly with $K$. TriviaQA is also the task in Table~\ref{tab:bench_main} where ASMI is not the dominant method.

\section{Boundary Deconfounds}
\label{app:boundary}
Two pre-registered checks ask whether the parametric-side boundary could be an artifact of the estimator rather than a property of the task regime.
\paragraph{Enlarged truncation.}
We re-ran the masked scoring on closed-book TriviaQA with the truncation enlarged from $K = 64$ to $K = 256$ ($n = 5\mathrm{K}$, identical per-example mask seeds). Mean token coverage rises to $0.97$, where the high-MI stratum of Table~\ref{tab:topk-highmi} is the stringent case rather than the population average. Sem-ASMI nonetheless stays significantly below MSP, a paired gap of $-0.041$ PRR (95\% CI $[-0.055, -0.025]$), and on the common subset the paired effect of enlarging $K$ is slightly negative ($-0.018$, CI $[-0.031, -0.005]$ for Sem-ASMI and $-0.026$, CI $[-0.046, -0.006]$ for raw MI). The parametric-side deficit is therefore not a truncation artifact. The adaptive variant was not re-scored, since its gate needs sample-embedding inputs not cached in this pipeline.

\paragraph{Same-dataset contrast.}
A second control holds the dataset fixed and moves only the knowledge source. We re-ran SQuAD on the same $5\mathrm{K}$ examples with the passage removed from the prompt, keeping questions, decoding, the AlignScore metric, and the ASMI operating point identical, and clustering both conditions by the original passage ($1{,}934$ clusters). The Sem-ASMI to MSP gap is a statistical tie in the open-book condition ($-0.034$, 95\% CI $[-0.078, +0.010]$) and significantly negative once the passage is removed ($-0.054$, CI $[-0.088, -0.021]$), which reproduces the parametric-side pattern within a single dataset. Removing the passage also raises the base error rate sharply (mean AlignScore $0.867$ to $0.109$), so we read the paired within-condition gaps. Absolute PRR levels are compressed at this error rate.

\section{The Distinct-Signal Analysis}
\label{app:distinct}
This appendix details the test behind the ``A distinct signal'' paragraph. Aggregate PRR and rank correlation cannot separate two questions: whether ASMI ranks errors well, and whether it reads information that output confidence does not. Because token-level $\mathrm{MI}_t$ and single-pass entropy are strongly rank correlated (Spearman around $0.9$ on CoQA), we judge distinctness at the source, by residualization, not by aggregate scores.
\paragraph{Setup.}
All quantities are computed at the operating layer. The confident-but-fragile and kernel analyses use Qwen3-4B-base, selecting BabiQA by the largest headroom-normalized gap between Sem-ASMI and MSP, with CoQA as a real-grounded generalization target and closed-book TriviaQA as the parametric contrast, and the residual test extends to all four datasets on both Qwen backbones (Table~\ref{tab:app-distinct}). For each generated sequence we record the semantically-weighted score $u^{\mathrm{sem}}$, the raw $\mathrm{MI}$, the single-pass entropy $H$, MSP, the ensemble-mean and mean-per-sample entropies used inside $\mathrm{MI}$, and the AlignScore label.
\paragraph{Residualization.}
We fit a logistic model of correctness on $\{\mathrm{MSP}, H\}$ and, out of fold, measure whether adding $u^{\mathrm{sem}}$ improves error prediction (incremental AUROC and a likelihood-ratio test) and whether the part of $u^{\mathrm{sem}}$ orthogonal to $\{\mathrm{MSP}, H\}$ detects errors on its own (residual AUROC against $0.5$). All splits are out of fold with a fixed seed and bootstrap confidence intervals. Across the grid the residual excludes chance in five of the six grounded cells, with CoQA on Qwen3-8B the one miss, and sits at chance on TriviaQA at both backbones (Table~\ref{tab:app-distinct}). The likelihood-ratio test is nominally significant on TriviaQA at Qwen3-8B, but both intervals include chance and the interval criterion is what we use throughout. The raw $\mathrm{MI}$ carries part of the same residual on BabiQA and is borderline on CoQA, and on BabiQA at Qwen3-8B it exceeds the weighted score ($0.687$ against $0.603$), so the kernel is not uniformly helpful at the larger scale.
\paragraph{SQuAD at Qwen3-4B.}
The two 5K rows are independent Monte Carlo draws of the same estimator, correlating at $0.95$ at the sequence level, and the effect sits at the estimator's resolution: one draw excludes chance and the other does not. Within a draw the signal strengthens under alternative aggregations, from $0.550$ for the mean to $0.594$ for the top-quartile mean and $0.613$ for the max, consistent with mean aggregation diluting a token-level signal over SQuAD's longest answers. This draw sensitivity concerns a borderline significance verdict rather than score instability, consistent with the redraw correlations of Appendix~\ref{app:stability}.
\begin{table*}[t]
\centering
\resizebox{0.65\linewidth}{!}{%
\begin{tabular}{llcc}
\toprule
Backbone & Dataset & Residual AUROC & Incr.\ AUROC \\
\midrule
Qwen3-4B & BabiQA & $0.589\,[0.547, 0.627]$ & $+0.020\,[+0.008, +0.032]$ \\
Qwen3-4B & CoQA & $0.549\,[0.501, 0.595]$ & $+0.018\,[+0.005, +0.031]$ \\
Qwen3-4B & SQuAD, analysis draw & $0.550\,[0.516, 0.584]$ & $+0.007\,[-0.000, +0.015]$ \\
Qwen3-4B & SQuAD, deployed draw & $0.501\,[0.466, 0.535]$ & $+0.001\,[-0.002, +0.003]$ \\
Qwen3-4B & TriviaQA & $0.495\,[0.465, 0.524]$ & $-0.001\,[-0.003, +0.000]$ \\\midrule
Qwen3-8B & BabiQA & $0.603\,[0.558, 0.641]$ & $+0.019\,[+0.009, +0.029]$ \\
Qwen3-8B & CoQA & $0.517\,[0.472, 0.560]$ & $+0.007\,[-0.000, +0.015]$ \\
Qwen3-8B & SQuAD & $0.543\,[0.504, 0.586]$ & $+0.004\,[-0.006, +0.014]$ \\
Qwen3-8B & TriviaQA & $0.511\,[0.475, 0.547]$ & $+0.002\,[-0.002, +0.006]$ \\
\bottomrule
\end{tabular}
}
\caption{Out-of-fold residual and incremental error-detection AUROC of $u^{\mathrm{sem}}$ over $\{\mathrm{MSP}, H\}$ (95\% CIs). A cell counts as detected when the residual interval excludes $0.5$. The two SQuAD Qwen3-4B rows are the same estimator under two independent Monte Carlo mask draws.}
\label{tab:app-distinct}
\end{table*}

\paragraph{The confident-but-fragile split.}
Splitting sequences at the medians of MSP and $u^{\mathrm{sem}}$ gives a two-by-two grid of error rates. On BabiQA the confident-fragile cell (high MSP, high $u^{\mathrm{sem}}$) carries an error gap of $0.21$ $[0.12, 0.29]$ over the confident-robust cell, the contrast plotted in Figure~\ref{fig:confident-fragile}. The gap is $0.12$ $[0.04, 0.21]$ on CoQA and $0.28$ $[0.19, 0.37]$ on TriviaQA, where a high base error rate inflates the absolute difference. The raw gap is therefore a poor regime marker on its own. The regimes separate only after conditioning on $\{\mathrm{MSP}, H\}$, where the grounded gaps still carry information and the TriviaQA gap does not (Table~\ref{tab:app-distinct}). Within the confident stratum the gap is not resolvable at the token level ($-0.09$ $[-0.16, 0.00]$), consistent with the sequence-level reading below.

\paragraph{What the kernel adds.}
The residual over $\{\mathrm{MSP}, H\}$ above tests distinctness from output confidence. A separate question is how much of the distinct signal comes from the semantic kernel rather than from the masking disagreement, and it cannot be read from the same residual. The two entropies inside the mutual information, the ensemble-mean entropy $H(\bar p_t)$ and the mean-per-sample entropy $\tfrac{1}{S}\sum_s H(p^{(s)}_t)$, span $\mathrm{MI}_t$ by construction, since $\mathrm{MI}_t$ is their difference (Eq.~\ref{eq:mi}). Controlling for both therefore removes the entire linear contribution of the mutual information, so the residual of $u^{\mathrm{sem}} = \mathrm{MI}_t(1 - A_t)$ under this control measures what the kernel factor $(1 - A_t)$ adds, not distinctness from confidence. So read, the kernel contributes error information beyond the masking disagreement on synthetic BabiQA ($0.603\,[0.565, 0.641]$) and little on real CoQA ($0.513\,[0.465, 0.562]$). This is not the distinct signal weakening on CoQA, which the $\{\mathrm{MSP}, H\}$ test in Table~\ref{tab:app-distinct} already rules out. It is that on real grounded data the distinct signal is carried by the attention-masking disagreement itself, and the kernel improves the ranking by discounting surface-form disagreement rather than by adding an axis orthogonal to $\mathrm{MI}_t$. This is consistent with the kernel's $0.04$ to $0.06$ PRR lift on CoQA in Table~\ref{tab:bench_main}, a reranking within the mutual-information ordering rather than new information. For the same span reason we do not control raw $\mathrm{MI}$ with these entropies, since its residual would be zero by construction.

\paragraph{Ensemble-mean entropy alone.}
A natural question is whether the masked ensemble helps only through averaging, since the ensemble-mean entropy $H(\bar p_t)$ is a free by-product of the same forty passes. Scored alone on the three cached Qwen3-4B analysis cells (BabiQA and CoQA at $n = 1\mathrm{K}$, TriviaQA at $n = 1.5\mathrm{K}$), it is essentially the single-pass entropy, with Spearman rank correlations from $0.967$ to $0.995$, so the ensemble mean adds no standalone information. Raw $\mathrm{MI}$ does not beat it on the grounded cells (both intervals cross zero), but the semantically weighted $u^{\mathrm{sem}}$ does, by $+0.046$ PRR on BabiQA (95\% CI $[+0.004, +0.089]$) and $+0.093$ on CoQA ($[+0.032, +0.156]$) under the same cluster-respecting paired bootstrap. On parametric TriviaQA the ordering reverses and $H(\bar p_t)$ alone is the best of the four, consistent with the regime boundary. Averaging over masks therefore accounts for none of the gain, which comes from the weighted disagreement term that the semantic kernel multiplies.

\paragraph{Further scope.}
Two limits of the analysis are worth recording. The distinctness is established by residualization and not by an asymmetric error set: at a matched flag rate on BabiQA, $u^{\mathrm{sem}}$ and entropy flag balanced disjoint error sets (McNemar $p = 0.26$), so the correct claim is that $u^{\mathrm{sem}}$ adds error-predictive information beyond output confidence, not that it catches a class of errors entropy is blind to. And within the confident stratum the effect is not resolvable at the token level with per-sequence labels (AUROC near $0.5$), so it is a sequence-level phenomenon.
\section{Deployment Payoff: Full Analysis}
\label{app:payoff}
This appendix gives the full selective-prediction analysis behind the deployment result in the main text. We report the single-signal risk-coverage comparison, the honest null for a global selector, and the parametric contrast, none of which fit in the main body.
\paragraph{Protocol.}
For each signal we sort predictions by uncertainty, retain the most confident fraction (the coverage $c$), and abstain on the rest. The retained error rate is the risk. We report the area under the risk-coverage curve (AURC, lower is better)~\citep{geifman2017selective}. Any selector that combines signals is fit out-of-fold with the same cluster keys as the main-text bootstrap, so no example scores itself. We report grounded cells (BabiQA, CoQA) and the parametric contrast (TriviaQA) throughout. SQuAD shows no confident-stratum gain on Qwen3-4B, consistent with the dilution pattern of Appendix~\ref{app:distinct}, so the payoff claim rests on BabiQA and CoQA.
\paragraph{Single-signal risk-coverage.}
Table~\ref{tab:payoff-aurc} reports AURC for each signal on its own. On both grounded cells $u^{\mathrm{sem}}$ has the lowest AURC, and on parametric TriviaQA it is the worst, mirroring the benchmark boundary. The advantage is concentrated at the most-confident end of the curve: $u^{\mathrm{sem}}$ is lowest at coverage $0.5$ to $0.6$ and crosses the single-pass signals near full coverage. This is the same confident-region effect that the confident-stratum filter exploits below, shown within the confident stratum in Figure~\ref{fig:payoff} of the main text.
\begin{table}[t]
\centering
\footnotesize
\setlength{\tabcolsep}{6pt}
\begin{tabular}{lccc}
\toprule
Cell & MSP & Entropy & $u^{\mathrm{sem}}$ \\
\midrule
BabiQA (grounded)     & $0.151$ & $0.151$ & $\mathbf{0.133}$ \\
CoQA (grounded)       & $0.134$ & $0.134$ & $\mathbf{0.123}$ \\
TriviaQA (parametric) & $0.314$ & $\mathbf{0.305}$ & $0.327$ \\
\bottomrule
\end{tabular}
\caption{Single-signal area under the risk-coverage curve (AURC, lower is better). Bold marks the best signal in each row.}
\label{tab:payoff-aurc}
\end{table}
\paragraph{A trained global selector.}
We next ask whether $u^{\mathrm{sem}}$ improves a trained selector over the single-pass signals. We fit an out-of-fold logistic model of correctness on $\{\mathrm{MSP}, \mathrm{entropy}\}$ (base) and on $\{\mathrm{MSP}, \mathrm{entropy}, u^{\mathrm{sem}}\}$ (full), and compare their risk-coverage curves. Across coverage $\{.7, .8, .9\}$ the two curves stay within noise on both grounded cells, with paired bootstrap intervals on the difference that include zero throughout, and likewise on TriviaQA. The reason is localization rather than absence: $u^{\mathrm{sem}}$'s error information sits almost entirely in the confident stratum, so a single global fit that spreads one model across all coverage levels averages that region together with the rest and cannot use it. The same information is what the confident-stratum filter below turns into a large gain.

\paragraph{Filtering the confident stratum.}
The payoff appears when the filter targets the confident stratum directly, which is the training-free result reported in the main text (Table~\ref{tab:payoff}, Figure~\ref{fig:payoff}). Within the MSP-confident half, abstaining on the top half by $u^{\mathrm{sem}}$ gives a final coverage of $0.25$ and roughly halves the retained error on BabiQA and CoQA, well ahead of spending the same abstention budget on predictive entropy. On parametric TriviaQA the effect reverses and entropy filtering is slightly better. The null global selector and the confident-stratum filter are consistent with each other, since the deployment value of $u^{\mathrm{sem}}$ is concentrated in the confident-but-fragile region the mechanism predicts.

\paragraph{Scope.}
The analysis covers two grounded cells and one parametric contrast on Qwen3-4B-base. $u^{\mathrm{sem}}$ runs on TriviaQA as well, with a finite AURC of $0.327$, and simply loses to the single-pass signals there rather than being inapplicable.
\section{Head-Level Structural Analysis}
\label{app:mechanism-details}
This appendix gives the head-level picture behind the ``Structural analysis'' paragraph, at the estimator's operating layer. It is deliberately brief, since the operating-layer validation of the signal is the confident-but-fragile analysis of Appendix~\ref{app:distinct}.
\paragraph{Estimation setup.}
The analysis uses Qwen3-4B-base with the same mask rate $p = 0.15$ and top-$K$ truncation $K = 64$ as the main experiments, at the operating depth of about $60\%$. The only change is an enlarged budget of $S_{\mathrm{an}} = 300$ masked passes per example to reduce the variance of the per-head estimates, leaving the estimator unchanged at $S = 40$. We analyze the first $500$ examples of each split, which yields $2{,}780$, $7{,}374$, $1{,}089$, and $2{,}027$ generated tokens on CoQA, SQuAD, BabiQA, and TriviaQA. Permutation nulls use $1$K shuffles per token and confidence intervals use $2$K example-level bootstrap resamples.
\paragraph{Criticality estimator.}
For each token $t$, the criticality of head $h$ is the difference in mean divergence between the masks that ablate it and those that keep it,
\begin{equation}
\hat{\beta}_{h,t} = \overline{d_t}\big|_{m^{(s)}_h = 0} - \overline{d_t}\big|_{m^{(s)}_h = 1},
\label{eq:criticality}
\end{equation}
where $\overline{d_t}\big|_{m^{(s)}_h = b}$ averages $d^{(s)}_t = \mathrm{KL}(p^{(s)}_t \| \bar{p}_t)$ over the samples with mask bit $b$. Because masks are independent Bernoulli variables, $\hat{\beta}_{h,t}$ is unbiased for the marginal effect of ablating head $h$. We measure the concentration of the non-negative profile by the Herfindahl index $\mathrm{HHI}_t = \sum_h w_{h,t}^2$ with $w_{h,t} = \max(\hat{\beta}_{h,t}, 0) / \sum_{h'} \max(\hat{\beta}_{h',t}, 0)$, ranging from $1/H$ for a uniform profile to $1$ for a single critical head. To remove the concentration that estimation noise alone produces, we permute the divergence-to-mask assignment $1$K times per token and standardize the observed index against this null, giving $z_t$.
\paragraph{Coupling at the operating layer.}
At the operating layer the criticality of an uncertain token is distributed across about ten heads rather than concentrated on a few (Table~\ref{tab:op-coupling}). The coupling is modest in magnitude but consistent in sign, following the grounded and parametric divide. The association $\rho(\mathrm{MI}_t, z_t)$ is positive on all three grounded benchmarks and negative on parametric TriviaQA, with intervals excluding zero, and mutual information tracks this head structure more than single-pass entropy does on the grounded cells. On CoQA the coupling sharpens with depth, from $0.09$ at the operating layer to $0.44$ four layers from the top, as the criticality concentrates onto fewer heads.
\begin{table*}[t]
\centering
\setlength{\tabcolsep}{5pt}
\begin{tabular}{lccc}
\toprule
Benchmark & $\rho(\mathrm{MI}_t, z_t)$ & $\rho(H_t, z_t)$ & $n_{\mathrm{eff}}$ (high MI) \\
\midrule
CoQA     & $0.09\ [0.05, 0.13]$    & $-0.04\ [-0.09, 0.00]$  & $9.7$ \\
SQuAD    & $0.14\ [0.11, 0.17]$    & $0.04\ [0.01, 0.07]$    & $9.1$ \\
BabiQA   & $0.14\ [0.08, 0.19]$    & $0.10\ [0.05, 0.15]$    & $9.6$ \\
TriviaQA & $-0.10\ [-0.14, -0.05]$ & $-0.19\ [-0.23, -0.14]$ & $9.6$ \\
\bottomrule
\end{tabular}
\caption{Head-criticality at the $60\%$ operating layer on Qwen3-4B-base (95\% CIs). $z_t$ is the permutation-standardized concentration of the head-criticality profile, and $n_{\mathrm{eff}} = 1 / \mathrm{HHI}_t$ is its effective head count on the high-MI stratum.}
\label{tab:op-coupling}
\end{table*}

\paragraph{Causal probe.}
A causal probe confirms that the fragility is head borne. For each benchmark we take the $200$ highest-MI and $200$ lowest-MI generated tokens and deterministically ablate the token's $k$ most critical heads. Ablating even one head flips a sizable fraction of the high-MI tokens while the low-MI tokens almost never flip, and the gap widens with $k$ up to $k = 8$ (Table~\ref{tab:causal-flip}). The dependence is present on all four benchmarks, so $\mathrm{MI}_t$ measures genuine reliance on specific heads. What follows the grounded and parametric divide is whether that reliance tracks uncertainty. For $k \geq 2$ the flip rates are in fact highest on TriviaQA, so fragility itself is strongest under parametric recall, and it is the alignment with correctness that is missing there. Mechanistic accounts of factual recall~\citep{geva2021transformer,dai2022knowledge,meng2022locating,geva2023dissecting} explain this case, since knowledge sits in mid-layer MLP sublayers while attention heads extract it, so head perturbation probes an extraction pathway that runs whether or not the underlying knowledge is strong.

\begin{table}[t]
\centering
\setlength{\tabcolsep}{6pt}
\resizebox{1.0\columnwidth}{!}{%
\begin{tabular}{lccccc}
\toprule
High-MI flip rate (\%) & $k{=}1$ & $k{=}2$ & $k{=}4$ & $k{=}8$ & $k{=}16$ \\
\midrule
CoQA     & $16$ & $33$ & $36$ & $43$ & $43$ \\
SQuAD    & $14$ & $19$ & $19$ & $25$ & $25$ \\
BabiQA   & $25$ & $32$ & $41$ & $42$ & $31$ \\
TriviaQA & $22$ & $35$ & $57$ & $63$ & $61$ \\
\bottomrule
\end{tabular}
}
\caption{Causal probe at the operating layer. Fraction of the $200$ highest-MI tokens whose greedy prediction flips when the token's $k$ most critical heads are ablated. The $200$ lowest-MI tokens flip at most $1\%$ throughout.}
\label{tab:causal-flip}
\end{table}
\section{Derivation of the Coverage-Fidelity Constraints}
\label{app:coverage-fidelity}
\begin{figure}[h]
\centering
\includegraphics[width=0.9\linewidth]{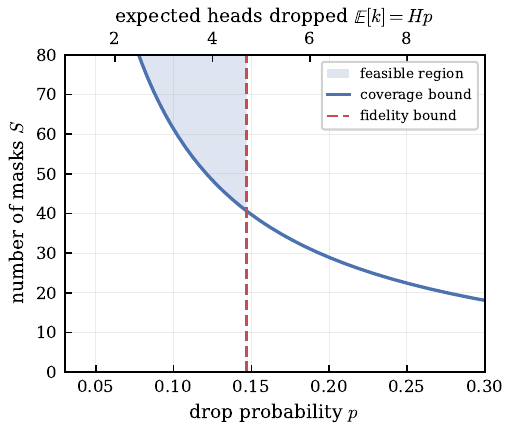}
\caption{Coverage lower bound and fidelity upper bound in the (p,S) plane. The adopted operating point (0.15, 40) is the minimum-cost corner of the feasible region.}
\label{fig:supp_fidelity}
\end{figure}
\begin{table*}[ht]
\centering
\resizebox{\linewidth}{!}{%
\begin{tabular}{l|cccccccc}
\toprule
\multirow{2}{*}{UE Method} & \multicolumn{2}{c}{Qwen3-4B} & \multicolumn{2}{c}{Qwen3-8B} & \multicolumn{2}{c}{Llama-2-7B} & \multicolumn{2}{c}{Mistral-7B} \\
\cline{2-9}
 & CoQA & SQuAD & CoQA & SQuAD & CoQA & SQuAD & CoQA & SQuAD \\
\midrule\midrule
\multicolumn{9}{l}{\textit{Information-based}} \\
Maximum Sequence Probability & 0.45$\pm$0.02 & 0.32$\pm$0.03 & 0.53$\pm$0.02 & 0.35$\pm$0.02 & 0.53$\pm$0.01 & 0.18$\pm$0.02 & 0.52$\pm$0.02 & 0.65$\pm$0.01 \\
Perplexity & 0.42$\pm$0.02 & 0.27$\pm$0.03 & 0.50$\pm$0.01 & 0.40$\pm$0.02 & 0.52$\pm$0.01 & \underline{0.73$\pm$0.02} & 0.48$\pm$0.02 & 0.81$\pm$0.01 \\
Mean Token Entropy & 0.40$\pm$0.02 & 0.27$\pm$0.03 & 0.50$\pm$0.01 & 0.41$\pm$0.02 & 0.46$\pm$0.02 & 0.68$\pm$0.02 & 0.44$\pm$0.02 & 0.78$\pm$0.01 \\
Pointwise Mutual Information & -0.05$\pm$0.02 & -0.14$\pm$0.04 & -0.04$\pm$0.02 & 0.24$\pm$0.02 & -0.04$\pm$0.02 & 0.50$\pm$0.02 & -0.12$\pm$0.03 & 0.69$\pm$0.02 \\
SelfCertainty & 0.19$\pm$0.02 & 0.02$\pm$0.04 & 0.28$\pm$0.02 & 0.32$\pm$0.02 & 0.21$\pm$0.02 & 0.57$\pm$0.02 & 0.23$\pm$0.02 & 0.73$\pm$0.02 \\
Conditional Pointwise MI & -0.19$\pm$0.03 & -0.35$\pm$0.05 & -0.22$\pm$0.02 & -0.15$\pm$0.03 & -0.21$\pm$0.02 & 0.41$\pm$0.02 & -0.29$\pm$0.03 & 0.67$\pm$0.02 \\
\midrule\midrule
\multicolumn{9}{l}{\textit{Sample-diversity}} \\
Monte Carlo Sequence Entropy & 0.43$\pm$0.02 & 0.27$\pm$0.03 & 0.52$\pm$0.02 & 0.45$\pm$0.02 & 0.52$\pm$0.01 & 0.36$\pm$0.02 & 0.51$\pm$0.02 & 0.77$\pm$0.01 \\
MC Normalized Sequence Entropy & 0.44$\pm$0.02 & 0.23$\pm$0.03 & 0.49$\pm$0.02 & 0.39$\pm$0.02 & 0.50$\pm$0.01 & 0.48$\pm$0.02 & 0.49$\pm$0.02 & 0.69$\pm$0.01 \\
Semantic Entropy & 0.46$\pm$0.02 & 0.28$\pm$0.03 & 0.53$\pm$0.01 & 0.45$\pm$0.02 & \underline{0.55$\pm$0.01} & 0.47$\pm$0.02 & \underline{0.54$\pm$0.02} & 0.80$\pm$0.01 \\
SAR & 0.46$\pm$0.02 & 0.27$\pm$0.03 & 0.47$\pm$0.02 & 0.40$\pm$0.02 & 0.53$\pm$0.01 & 0.51$\pm$0.02 & 0.52$\pm$0.02 & 0.78$\pm$0.01 \\
SentenceSAR (max 5) & 0.44$\pm$0.02 & 0.28$\pm$0.03 & 0.51$\pm$0.02 & 0.43$\pm$0.02 & 0.54$\pm$0.01 & 0.45$\pm$0.02 & 0.53$\pm$0.02 & 0.80$\pm$0.01 \\
LUQ & 0.37$\pm$0.02 & 0.32$\pm$0.02 & 0.37$\pm$0.02 & \textbf{0.50$\pm$0.02} & 0.39$\pm$0.02 & 0.43$\pm$0.02 & 0.41$\pm$0.02 & 0.73$\pm$0.01 \\
Kernel Language Entropy & 0.41$\pm$0.02 & 0.31$\pm$0.02 & 0.40$\pm$0.02 & \underline{0.48$\pm$0.02} & 0.44$\pm$0.02 & 0.50$\pm$0.02 & 0.47$\pm$0.02 & 0.80$\pm$0.01 \\
\midrule\midrule
\multicolumn{9}{l}{\textit{Probing}} \\
P(True) & 0.03$\pm$0.02 & 0.07$\pm$0.03 & 0.07$\pm$0.02 & 0.17$\pm$0.02 & -0.04$\pm$0.02 & 0.47$\pm$0.02 & -0.05$\pm$0.03 & 0.78$\pm$0.01 \\
P(True) Sampling & -0.04$\pm$0.03 & 0.03$\pm$0.04 & 0.04$\pm$0.02 & 0.30$\pm$0.02 & -0.04$\pm$0.02 & 0.44$\pm$0.02 & -0.28$\pm$0.03 & 0.29$\pm$0.03 \\
\midrule\midrule
\multicolumn{9}{l}{\textit{Attention-based}} \\
RAUQ & 0.37$\pm$0.02 & \underline{0.32$\pm$0.03} & 0.48$\pm$0.02 & 0.45$\pm$0.02 & 0.47$\pm$0.02 & 0.46$\pm$0.02 & 0.46$\pm$0.02 & 0.82$\pm$0.01 \\
RAUQ (entropy) & 0.47$\pm$0.02 & \textbf{0.36$\pm$0.02} & 0.53$\pm$0.02 & 0.47$\pm$0.02 & 0.50$\pm$0.02 & \textbf{0.75$\pm$0.02} & 0.51$\pm$0.02 & \underline{0.83$\pm$0.01} \\
\midrule\midrule
\multicolumn{9}{l}{\textit{Attention-based, active perturbation (ours)}} \\
ASMI (60\%, K=64) & 0.46$\pm$0.02 & 0.29$\pm$0.03 & 0.51$\pm$0.01 & 0.34$\pm$0.02 & 0.49$\pm$0.02 & 0.62$\pm$0.02 & 0.48$\pm$0.02 & 0.77$\pm$0.02 \\
Sem-ASMI (60\%, K=64) & \underline{0.52$\pm$0.02} & 0.29$\pm$0.03 & \underline{0.55$\pm$0.01} & 0.40$\pm$0.02 & 0.54$\pm$0.01 & 0.69$\pm$0.02 & 0.53$\pm$0.02 & 0.83$\pm$0.01 \\
Adapt-ASMI (60\%, K=64) & \textbf{0.53$\pm$0.01} & 0.31$\pm$0.02 & \textbf{0.56$\pm$0.02} & 0.40$\pm$0.02 & \textbf{0.55$\pm$0.01} & 0.71$\pm$0.02 & \textbf{0.54$\pm$0.02} & \textbf{0.84$\pm$0.01} \\\midrule
ASMI (70\%, K=64) & 0.47$\pm$0.02 & 0.27$\pm$0.03 & 0.52$\pm$0.01 & 0.40$\pm$0.02 & 0.46$\pm$0.02 & 0.61$\pm$0.02 & 0.46$\pm$0.02 & 0.73$\pm$0.02 \\
ASMI (80\%, K=64) & 0.44$\pm$0.02 & 0.28$\pm$0.02 & 0.51$\pm$0.01 & 0.43$\pm$0.02 & 0.40$\pm$0.02 & 0.67$\pm$0.02 & 0.38$\pm$0.02 & 0.74$\pm$0.02 \\
ASMI (90\%, K=64) & 0.29$\pm$0.02 & 0.24$\pm$0.03 & 0.42$\pm$0.02 & 0.34$\pm$0.02 & 0.43$\pm$0.02 & 0.70$\pm$0.01 & 0.45$\pm$0.02 & 0.73$\pm$0.02 \\
Sem-ASMI (70\%, K=64) & 0.52$\pm$0.02 & 0.28$\pm$0.03 & 0.54$\pm$0.01 & 0.42$\pm$0.02 & 0.52$\pm$0.01 & 0.68$\pm$0.02 & 0.53$\pm$0.02 & 0.79$\pm$0.01 \\
Sem-ASMI (80\%, K=64) & 0.50$\pm$0.01 & 0.28$\pm$0.03 & 0.54$\pm$0.01 & 0.43$\pm$0.02 & 0.49$\pm$0.01 & 0.71$\pm$0.01 & 0.49$\pm$0.02 & 0.80$\pm$0.01 \\
Sem-ASMI (90\%, K=64) & 0.41$\pm$0.02 & 0.26$\pm$0.03 & 0.51$\pm$0.02 & 0.39$\pm$0.02 & 0.52$\pm$0.01 & 0.74$\pm$0.01 & 0.52$\pm$0.02 & 0.78$\pm$0.01 \\
Adapt-ASMI (70\%, K=64) & 0.52$\pm$0.02 & 0.29$\pm$0.03 & 0.55$\pm$0.01 & 0.44$\pm$0.02 & 0.53$\pm$0.01 & 0.69$\pm$0.02 & 0.54$\pm$0.01 & 0.81$\pm$0.01 \\
Adapt-ASMI (80\%, K=64) & 0.50$\pm$0.02 & 0.30$\pm$0.02 & 0.55$\pm$0.01 & 0.46$\pm$0.02 & 0.50$\pm$0.01 & 0.72$\pm$0.01 & 0.50$\pm$0.02 & 0.81$\pm$0.01 \\
Adapt-ASMI (90\%, K=64) & 0.42$\pm$0.02 & 0.27$\pm$0.03 & 0.52$\pm$0.02 & 0.40$\pm$0.02 & 0.53$\pm$0.01 & 0.75$\pm$0.01 & 0.53$\pm$0.02 & 0.80$\pm$0.01 \\
\bottomrule
\end{tabular}
}
\caption{PRR (mean $\pm$ standard error, higher is better) on CoQA and SQuAD, scored by AlignScore. Full version of Table~\ref{tab:bench_main}, with all 17 baselines and the three ASMI variants across the depth sweep. \textbf{Bold} and \underline{underline} mark the best and second-best PRR per column at the operating depth $d = 60\%$. Rows at other depths belong to the sweep and are left unmarked.}
\label{tab:main-ours-coqa-squad}
\end{table*}
\begin{table*}[ht!]
\centering
\resizebox{\linewidth}{!}{%
\begin{tabular}{l|cccccccc}
\toprule
\multirow{2}{*}{UE Method} & \multicolumn{2}{c}{Qwen3-4B} & \multicolumn{2}{c}{Qwen3-8B} & \multicolumn{2}{c}{Llama-2-7B} & \multicolumn{2}{c}{Mistral-7B} \\
\cline{2-9}
 & BabiQA & TriviaQA & BabiQA & TriviaQA & BabiQA & TriviaQA & BabiQA & TriviaQA \\
\midrule\midrule
\multicolumn{9}{l}{\textit{Information-based}} \\
Maximum Sequence Probability & 0.48$\pm$0.04 & \underline{0.72$\pm$0.02} & 0.76$\pm$0.03 & \underline{0.70$\pm$0.02} & \textbf{0.58$\pm$0.04} & 0.77$\pm$0.01 & 0.59$\pm$0.04 & 0.78$\pm$0.01 \\
Perplexity & 0.49$\pm$0.05 & 0.71$\pm$0.01 & 0.74$\pm$0.03 & 0.68$\pm$0.02 & 0.53$\pm$0.04 & 0.76$\pm$0.01 & 0.59$\pm$0.04 & \underline{0.79$\pm$0.01} \\
Mean Token Entropy & 0.56$\pm$0.04 & 0.70$\pm$0.02 & 0.75$\pm$0.03 & 0.69$\pm$0.02 & 0.30$\pm$0.05 & 0.75$\pm$0.01 & 0.54$\pm$0.04 & 0.78$\pm$0.01 \\
Pointwise Mutual Information & 0.24$\pm$0.05 & -0.05$\pm$0.02 & 0.46$\pm$0.04 & -0.07$\pm$0.02 & 0.20$\pm$0.05 & -0.02$\pm$0.03 & 0.24$\pm$0.05 & -0.01$\pm$0.03 \\
SelfCertainty & 0.07$\pm$0.07 & 0.32$\pm$0.02 & 0.62$\pm$0.04 & 0.34$\pm$0.02 & 0.06$\pm$0.05 & 0.52$\pm$0.02 & 0.20$\pm$0.05 & 0.60$\pm$0.02 \\
Conditional Pointwise MI & 0.09$\pm$0.07 & -0.30$\pm$0.02 & -0.45$\pm$0.06 & -0.35$\pm$0.02 & 0.25$\pm$0.05 & -0.34$\pm$0.03 & 0.09$\pm$0.06 & -0.35$\pm$0.03 \\
\midrule\midrule
\multicolumn{9}{l}{\textit{Sample-diversity}} \\
Monte Carlo Sequence Entropy & 0.47$\pm$0.05 & 0.71$\pm$0.02 & 0.74$\pm$0.03 & 0.69$\pm$0.01 & 0.52$\pm$0.04 & 0.76$\pm$0.01 & 0.56$\pm$0.04 & 0.77$\pm$0.01 \\
MC Normalized Sequence Entropy & 0.50$\pm$0.05 & 0.70$\pm$0.02 & 0.72$\pm$0.03 & 0.67$\pm$0.02 & 0.50$\pm$0.04 & 0.75$\pm$0.01 & 0.53$\pm$0.04 & 0.78$\pm$0.01 \\
Semantic Entropy & 0.54$\pm$0.04 & 0.72$\pm$0.02 & 0.74$\pm$0.03 & 0.70$\pm$0.02 & 0.50$\pm$0.04 & \underline{0.77$\pm$0.01} & 0.58$\pm$0.04 & 0.78$\pm$0.01 \\
SAR & \underline{0.63$\pm$0.04} & 0.71$\pm$0.02 & 0.71$\pm$0.03 & 0.68$\pm$0.02 & 0.49$\pm$0.04 & \textbf{0.78$\pm$0.01} & 0.56$\pm$0.04 & \textbf{0.80$\pm$0.01} \\
SentenceSAR (max 5) & 0.49$\pm$0.05 & 0.71$\pm$0.02 & 0.73$\pm$0.03 & 0.69$\pm$0.02 & 0.52$\pm$0.04 & 0.77$\pm$0.01 & 0.54$\pm$0.04 & 0.78$\pm$0.01 \\
LUQ & 0.47$\pm$0.05 & 0.64$\pm$0.02 & 0.43$\pm$0.05 & 0.52$\pm$0.02 & 0.42$\pm$0.04 & 0.69$\pm$0.02 & 0.37$\pm$0.05 & 0.73$\pm$0.02 \\
Kernel Language Entropy & 0.48$\pm$0.05 & 0.64$\pm$0.02 & 0.44$\pm$0.05 & 0.51$\pm$0.02 & 0.41$\pm$0.05 & 0.70$\pm$0.02 & 0.37$\pm$0.05 & 0.73$\pm$0.02 \\
\midrule\midrule
\multicolumn{9}{l}{\textit{Probing}} \\
P(True) & 0.49$\pm$0.04 & 0.30$\pm$0.02 & 0.22$\pm$0.06 & 0.39$\pm$0.02 & 0.21$\pm$0.05 & 0.06$\pm$0.03 & 0.20$\pm$0.05 & -0.11$\pm$0.03 \\
P(True) Sampling & 0.11$\pm$0.06 & 0.38$\pm$0.02 & 0.32$\pm$0.06 & 0.39$\pm$0.02 & 0.21$\pm$0.05 & 0.06$\pm$0.03 & 0.18$\pm$0.06 & 0.20$\pm$0.03 \\
\midrule\midrule
\multicolumn{9}{l}{\textit{Attention-based}} \\
RAUQ & 0.32$\pm$0.06 & 0.70$\pm$0.02 & 0.74$\pm$0.03 & 0.66$\pm$0.02 & \underline{0.57$\pm$0.04} & 0.75$\pm$0.01 & 0.48$\pm$0.05 & 0.70$\pm$0.02 \\
RAUQ (entropy) & 0.42$\pm$0.05 & \textbf{0.72$\pm$0.02} & 0.74$\pm$0.03 & \textbf{0.70$\pm$0.02} & 0.29$\pm$0.05 & 0.77$\pm$0.01 & 0.46$\pm$0.04 & 0.77$\pm$0.01 \\
\midrule\midrule
\multicolumn{9}{l}{\textit{Attention-based, active perturbation (ours)}} \\
ASMI (60\%, K=64) & 0.60$\pm$0.04 & 0.59$\pm$0.02 & \textbf{0.80$\pm$0.03} & 0.62$\pm$0.02 & 0.11$\pm$0.05 & 0.66$\pm$0.02 & 0.52$\pm$0.04 & 0.74$\pm$0.01 \\
Sem-ASMI (60\%, K=64) & \textbf{0.66$\pm$0.04} & 0.66$\pm$0.02 & 0.78$\pm$0.03 & 0.66$\pm$0.02 & 0.27$\pm$0.05 & 0.71$\pm$0.01 & \underline{0.61$\pm$0.04} & 0.78$\pm$0.01 \\
Adapt-ASMI (60\%, K=64) & \textbf{0.66$\pm$0.04} & 0.66$\pm$0.02 & \underline{0.79$\pm$0.03} & 0.67$\pm$0.02 & 0.29$\pm$0.05 & 0.72$\pm$0.01 & \textbf{0.61$\pm$0.03} & 0.78$\pm$0.01 \\\midrule
ASMI (70\%, K=64) & 0.48$\pm$0.04 & 0.64$\pm$0.02 & 0.78$\pm$0.03 & 0.65$\pm$0.02 & 0.23$\pm$0.05 & 0.69$\pm$0.01 & 0.41$\pm$0.04 & 0.67$\pm$0.01 \\
ASMI (80\%, K=64) & 0.33$\pm$0.04 & 0.58$\pm$0.02 & 0.48$\pm$0.04 & 0.62$\pm$0.02 & 0.24$\pm$0.05 & 0.66$\pm$0.02 & 0.51$\pm$0.04 & 0.68$\pm$0.01 \\
ASMI (90\%, K=64) & 0.55$\pm$0.04 & 0.58$\pm$0.02 & 0.76$\pm$0.03 & 0.49$\pm$0.02 & 0.10$\pm$0.04 & 0.68$\pm$0.02 & 0.49$\pm$0.04 & 0.69$\pm$0.02 \\
Sem-ASMI (70\%, K=64) & 0.58$\pm$0.04 & 0.68$\pm$0.02 & 0.76$\pm$0.03 & 0.68$\pm$0.02 & 0.33$\pm$0.04 & 0.73$\pm$0.01 & 0.57$\pm$0.04 & 0.74$\pm$0.01 \\
Sem-ASMI (80\%, K=64) & 0.38$\pm$0.05 & 0.64$\pm$0.02 & 0.70$\pm$0.03 & 0.66$\pm$0.02 & 0.39$\pm$0.04 & 0.72$\pm$0.01 & 0.63$\pm$0.03 & 0.75$\pm$0.01 \\
Sem-ASMI (90\%, K=64) & 0.64$\pm$0.04 & 0.67$\pm$0.02 & 0.77$\pm$0.03 & 0.62$\pm$0.02 & 0.31$\pm$0.05 & 0.73$\pm$0.01 & 0.62$\pm$0.03 & 0.75$\pm$0.01 \\
Adapt-ASMI (70\%, K=64) & 0.58$\pm$0.04 & 0.69$\pm$0.02 & 0.77$\pm$0.03 & 0.68$\pm$0.02 & 0.34$\pm$0.04 & 0.74$\pm$0.01 & 0.57$\pm$0.04 & 0.75$\pm$0.01 \\
Adapt-ASMI (80\%, K=64) & 0.38$\pm$0.05 & 0.64$\pm$0.02 & 0.69$\pm$0.03 & 0.66$\pm$0.02 & 0.41$\pm$0.04 & 0.72$\pm$0.01 & 0.63$\pm$0.03 & 0.75$\pm$0.01 \\
Adapt-ASMI (90\%, K=64) & 0.64$\pm$0.04 & 0.67$\pm$0.02 & 0.77$\pm$0.03 & 0.62$\pm$0.02 & 0.33$\pm$0.05 & 0.73$\pm$0.01 & 0.62$\pm$0.03 & 0.75$\pm$0.01 \\
\bottomrule
\end{tabular}
}
\caption{PRR (mean $\pm$ standard error, higher is better) on BabiQA and closed-book TriviaQA, the parametric control, scored by AlignScore. Full version of Table~\ref{tab:bench_main}, with all 17 baselines and the three ASMI variants across the depth sweep. \textbf{Bold} and \underline{underline} mark the best and second-best PRR per column at the operating depth $d = 60\%$. Rows at other depths belong to the sweep and are left unmarked.}
\label{tab:main-ours-babi-trivia}
\end{table*}
\paragraph{Coverage lower bound on $S$.}
Each Monte Carlo sample draws an independent mask $m^{(s)} \in \{0,1\}^H$ with $m^{(s)}_h \sim \mathrm{Bernoulli}(1-p)$. The probability that a given head $h$ is never masked across $S$ samples is $(1-p)^S$. By a union bound over $H$ heads, the probability that at least one head is never ablated is at most $H(1-p)^S$. Requiring this failure probability to be at most $\delta$ gives
\begin{equation}
H(1-p)^S \leq \delta
\;\;\Longleftrightarrow\;\;
S \geq \frac{\ln(H/\delta)}{-\ln(1-p)}.
\label{eq:app-coverage}
\end{equation}
Coverage is a necessary condition for the mutual-information estimate: a head that is never masked contributes no observed perturbation, so its criticality is invisible to the estimator. For $H{=}32$ and $\delta{=}0.05$, Equation~\ref{eq:app-coverage} yields $S \geq 62$ at $p{=}0.10$ but only $S \geq 40$ at $p{=}0.15$: increasing $p$ reduces the sampling budget required for full coverage.
\paragraph{Fidelity upper bound on $p$.}
The number of simultaneously masked heads in a sample is $k \sim \mathrm{Binomial}(H, p)$, with mean $Hp$ and variance $Hp(1-p)$. The perturbed network must remain a functioning model: if too many heads are removed at once, the measurement no longer probes dependence on attention paths but instead reflects generic degradation of a broken computation. We therefore require that $k$ stay below a tolerance $k_{\max}$ with high probability, using the Gaussian upper bound
\begin{equation}
Hp + z_{1-\delta'}\sqrt{Hp(1-p)} \;\leq\; k_{\max}.
\label{eq:app-fidelity}
\end{equation}
We set $k_{\max}{=}8$, i.e., one quarter of the heads in a layer, following head-pruning studies that report negligible loss within this margin and a sharp collapse beyond it~\citep{michel2019sixteen,voita2019analyzing}. With $z_{0.95}{=}1.645$ and $H{=}32$, the left side of Equation~\ref{eq:app-fidelity} evaluates to $4.8 + 1.645\sqrt{4.08} \approx 8.1$ at $p{=}0.15$, marginally above $k_{\max}$ under the Gaussian surrogate, while the exact $\mathrm{Binomial}(32, 0.15)$ 95th percentile equals $8$. The exact distribution therefore admits $p{=}0.15$ as the largest admissible mask rate, which is the sense in which we write $p \lesssim 0.15$.

\paragraph{Feasible region and the chosen operating point.}
Figure~\ref{fig:supp_fidelity} plots both constraints in the $(p, S)$ plane. The coverage curve (Eq.~\ref{eq:app-coverage}) bounds the region from below and decreases in $p$, and the fidelity line (Eq.~\ref{eq:app-fidelity}) bounds it from the right. The minimum-cost point of the feasible region, the smallest $S$ satisfying coverage at the largest admissible $p$, is the corner $(p, S) = (0.15, 40)$, which we adopt without further tuning. Earlier configurations $(0.10, 16)$ and $(0.15, 32)$ lie below the coverage curve. At $(0.10, 16)$ the expected number of never-ablated heads is $H(1-p)^S = 32 \cdot 0.9^{16} \approx 5.9$, so roughly six heads per layer are never perturbed across the $S$ samples and part of the signal is omitted.

\paragraph{Scope.}
Both constraints depend on the model only through the number of maskable units, which in our design is the number of query heads $H$, since masks are applied to query-head outputs before the output projection. Grouped-query attention reduces the number of key/value projections but leaves the query heads intact: Qwen3-4B, Qwen3-8B, and Mistral-7B retain $H = 32$ query heads alongside their grouped key/value projections, matching the standard multi-head attention of Llama-2-7B, so the operating point $(0.15, 40)$ transfers across all four backbones without re-derivation. Equations~\ref{eq:app-coverage} and~\ref{eq:app-fidelity} generalize directly, and re-derivation is required only when $H$ itself changes, for instance for a backbone with a different query-head count or a coarser masking granularity such as whole key/value groups. The coverage condition guarantees only that every head is perturbed at least once. It does not bound the variance of the MI estimate, which continues to decrease as $O(1/\sqrt{S})$ beyond the coverage threshold. Pairwise co-ablation coverage, in which every head \emph{pair} is observed masked jointly, would require $S \gtrsim \ln\!\big(\tbinom{H}{2}/\delta\big)/p^2 \approx 410$ samples at $p{=}0.15$ (at the same coverage failure probability $\delta{=}0.05$) and is outside our compute budget, so the estimator captures marginal head criticality rather than higher-order interaction effects.
\section{Measured Cost}
\label{app:cost}

All measurements use Qwen3-4B-base on a single H100 MIG 3g.40gb slice with 200 fixed examples per dataset, 10 warmup examples excluded, reporting per-example medians. Peak memory spans $8.3$ to $9.3$ GB across all methods. Phase medians on CoQA and SQuAD are: greedy response $4.14$ and $0.75$ seconds, ten stochastic samples drawn as one batch $4.41$ and $1.28$, NLI clustering for Semantic Entropy $0.18$ and $0.13$, mask-batched ASMI scoring $2.21$ and $1.74$, sequential ASMI scoring $5.17$ and $2.34$. Mask batching carries the comparison. The $40$ masked suffixes share one batched pass, which is $2.3\times$ faster than sequential masking and, on CoQA, cheaper than drawing the ten stochastic samples it replaces. The auxiliary NLI model contributes under $0.2$ seconds per example, so the cost of the sampling baselines is dominated by generation itself. The advantage of ASMI tracks input length and therefore favors the long CoQA inputs over the shorter SQuAD ones. The analytic FLOPs view of Appendix~\ref{app:impl-details} is complementary, since batched masked passes trade FLOPs for parallelism. Absolute seconds are specific to this hardware, so we read only the relative ordering from them.

\section{Full Benchmark and Depth Sweep}
\label{app:full-benchmark}
\label{app:full-sweep}
Tables~\ref{tab:main-ours-coqa-squad} and~\ref{tab:main-ours-babi-trivia} report the complete benchmark that the main text abbreviates. The main comparison in Table~\ref{tab:bench_main} keeps a compact baseline set at the single operating depth $d = 60\%$. Here we list every baseline together with the three ASMI variants across the full depth sweep, so the abbreviated main table and the depth analysis draw on one set of numbers. Bold and underline follow the main-table convention and are computed only at the $60\%$ operating depth, so the highlighting stays comparable to Table~\ref{tab:bench_main} while the remaining depths are shown for context. One property of the Llama-2-7B SQuAD column is worth noting. On this task $28.5\%$ of the greedy generations contain nothing but whitespace and wrong answers are much shorter than correct ones (7.0 against 12.7 tokens on average), which depresses length-sensitive scores such as MSP in that column, whose error-detection AUROC falls to $0.47$. Since every method scores the same greedy responses, comparisons within the column remain valid.

\paragraph{Depth selection.} We take $d = 60\%$ as the representative depth by aggregate PRR over the sweep. Aggregated across all sixteen cells, four benchmarks by four backbones, normalized PRR is highest at $60\%$ ($0.600$) against $0.592$, $0.577$, and $0.586$ at the deeper settings, and $60\%$ is the best or tied-best depth in eight of the twelve grounded columns. The exceptions are SQuAD on Qwen3-8B and Llama-2-7B and BabiQA on Llama-2-7B and Mistral-7B. On CoQA performance rises smoothly toward $60\%$, consistent with masking within the context-routing band rather than the near-output layers, and SQuAD follows the same trend on Qwen3-4B and Mistral-7B while favoring deeper masking on Llama-2-7B. BabiQA is the exception discussed next.

\paragraph{Stability of depth selection}
We verify that the $60\%$ choice is not an artifact of the full-data aggregate with a cluster-respecting split-half test. We draw $R = 1\mathrm{K}$ random halves (seed $20260711$), keeping related items together: CoQA is split by story and SQuAD by passage, the same cluster keys used for the main-text bootstrap, while BabiQA and TriviaQA are split by input. On each split we select the argmax depth on the first half and measure its out-of-half regret, the PRR gap to the best depth, on the second half, with PRR normalized within each half. Across the sixteen cells, $60\%$ is selected in $99.3\%$ of the splits, the remaining $0.7\%$ go to $70\%$, and $80\%$ and $90\%$ are never selected, with zero out-of-half regret in $98.7\%$ of splits and a mean regret of $0.0001$ PRR. The selection reproduces for Sem-ASMI at $99.0\%$. The representative depth is therefore stable across resampling and variant.
\begin{table}[t]
\centering
\footnotesize
\setlength{\tabcolsep}{5pt}
\begin{tabular}{lccccc}
\toprule
BabiQA cell & $\overline{\mathrm{MI}}$ & $\mathrm{MI}_{\mathrm{corr}}$ & $\mathrm{MI}_{\mathrm{wrong}}$ & ASMI & MSP \\
\midrule
Qwen3-4B   & $0.016$ & $0.014$ & $0.022$ & $0.77$ & $0.68$ \\
Qwen3-8B   & $0.025$ & $0.017$ & $0.045$ & $0.86$ & $0.86$ \\
Llama-2-7B & $0.002$ & $0.002$ & $0.002$ & $0.60$ & $0.75$ \\
\bottomrule
\end{tabular}
\caption{Mechanism of the BabiQA failure ($d = 60\%$, $K = 64$). Mean token-level $\mathrm{MI}_t$ split by answer correctness, with error-detection AUROC for ASMI and MSP. Head masking barely moves the Llama-2-7B output, which leaves correct and wrong answers indistinguishable to ASMI.}
\label{tab:babiqa-mechanism}
\end{table}
\begin{figure}[ht!]
\centering
\includegraphics[width=\columnwidth]{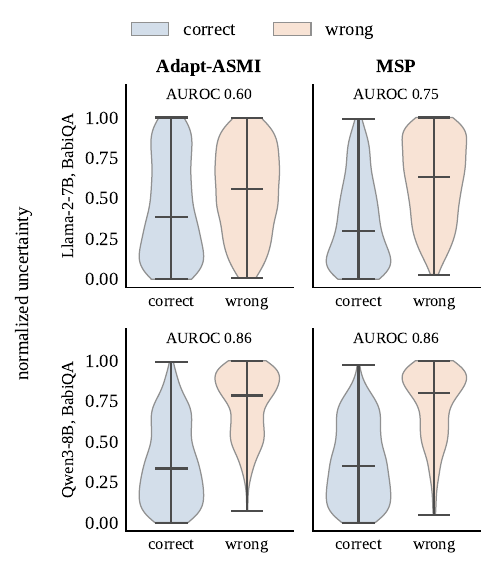}
\caption{Error separation on BabiQA. Violins show per-example uncertainty for correct (blue) and wrong (orange) generations, with error-detection AUROC per panel. Llama-2-7B (top) is over-robust to head masking, so Adapt-ASMI cannot separate errors while MSP still can. Qwen3-8B (bottom) routes wrong answers through more fragile paths.}
\label{fig:error-separation}
\end{figure}
\begin{table*}[ht!]
\centering
\setlength{\tabcolsep}{5pt}
\resizebox{\linewidth}{!}{%
\begin{tabular}{llllccc}
\toprule
Backbone & Dataset & Top ASMI variant (PRR) & Best baseline (PRR) & Top $\Delta$PRR vs SemEnt & Top $\Delta$PRR vs best baseline & Sem-ASMI $\Delta$PRR vs SemEnt \\
\midrule\midrule
\multirow{4}{*}{Qwen3-4B}
 & CoQA     & Adapt-ASMI (0.524) & RAUQ-E (0.469) & $+0.065^{*}\ [+0.042, +0.088]$ & $+0.056^{*}\ [+0.035, +0.076]$ & $+0.060^{*}\ [+0.036, +0.085]$ \\
 & SQuAD    & Adapt-ASMI (0.305) & RAUQ-E (0.359) & $+0.024\ [-0.011, +0.059]$ & $-0.053^{*}\ [-0.098, -0.006]$ & $+0.004\ [-0.037, +0.044]$ \\
 & BabiQA   & Sem-ASMI (0.657)   & SAR (0.628)    & $+0.120^{*}\ [+0.051, +0.192]$ & $+0.030\ [-0.014, +0.072]$ & $+0.120^{*}\ [+0.051, +0.192]$ \\
 & TriviaQA & Adapt-ASMI (0.662) & RAUQ-E (0.723) & $-0.057^{*}\ [-0.072, -0.043]$ & $-0.060^{*}\ [-0.073, -0.048]$ & $-0.061^{*}\ [-0.076, -0.046]$ \\
\midrule
\multirow{4}{*}{Qwen3-8B}
 & CoQA     & Adapt-ASMI (0.559) & SemEnt (0.529) & $+0.029^{*}\ [+0.013, +0.045]$ & $+0.029^{*}\ [+0.013, +0.045]$ & $+0.019^{*}\ [+0.002, +0.035]$ \\
 & SQuAD    & Adapt-ASMI (0.402) & LUQ (0.500)    & $-0.052^{*}\ [-0.087, -0.019]$ & $-0.098^{*}\ [-0.138, -0.059]$ & $-0.057^{*}\ [-0.094, -0.021]$ \\
 & BabiQA   & ASMI (0.798)       & MSP (0.761)    & $+0.060^{*}\ [+0.028, +0.095]$ & $+0.037^{*}\ [+0.005, +0.072]$ & $+0.042^{*}\ [+0.014, +0.074]$ \\
 & TriviaQA & Adapt-ASMI (0.670) & RAUQ-E (0.706) & $-0.031^{*}\ [-0.044, -0.018]$ & $-0.037^{*}\ [-0.051, -0.022]$ & $-0.037^{*}\ [-0.051, -0.023]$ \\
\midrule
\multirow{4}{*}{Llama-2-7B}
 & CoQA     & Adapt-ASMI (0.552) & SemEnt (0.545) & $+0.008\ [-0.012, +0.027]$ & $+0.008\ [-0.012, +0.027]$ & $-0.001\ [-0.023, +0.020]$ \\
 & SQuAD & Adapt-ASMI (0.706) & RAUQ-E (0.753) & $+0.236^{*}\ [+0.202, +0.270]$ & $-0.047^{*}\ [-0.062, -0.033]$ & $+0.224^{*}\ [+0.187, +0.261]$ \\
 & BabiQA   & Adapt-ASMI (0.286) & MSP (0.581)    & $-0.215^{*}\ [-0.283, -0.148]$ & $-0.295^{*}\ [-0.365, -0.228]$ & $-0.227^{*}\ [-0.300, -0.157]$ \\
 & TriviaQA & Adapt-ASMI (0.723) & RAUQ-E (0.772) & $-0.048^{*}\ [-0.066, -0.030]$ & $-0.049^{*}\ [-0.064, -0.033]$ & $-0.056^{*}\ [-0.074, -0.037]$ \\
\midrule
\multirow{4}{*}{Mistral-7B}
 & CoQA     & Adapt-ASMI (0.537) & SemEnt (0.536) & $+0.002\ [-0.018, +0.021]$ & $+0.002\ [-0.018, +0.021]$ & $-0.007\ [-0.028, +0.014]$ \\
 & SQuAD    & Adapt-ASMI (0.840) & RAUQ-E (0.830) & $+0.037^{*}\ [+0.015, +0.059]$ & $+0.010\ [-0.003, +0.023]$ & $+0.025\ [-0.001, +0.051]$ \\
 & BabiQA   & Adapt-ASMI (0.607) & MSP (0.593)    & $+0.024\ [-0.029, +0.077]$ & $+0.015\ [-0.032, +0.062]$ & $+0.022\ [-0.032, +0.077]$ \\
 & TriviaQA & Adapt-ASMI (0.784) & SAR (0.795)    & $+0.000\ [-0.013, +0.014]$ & $-0.012\ [-0.026, +0.003]$ & $-0.002\ [-0.016, +0.012]$ \\
\bottomrule
\end{tabular}
}
\caption{Head-to-head paired significance at the $60\%$ operating depth. Each entry is a paired $\Delta$PRR with a 95\% percentile interval from a cluster-respecting paired bootstrap ($B = 10\mathrm{K}$, using the cluster keys and seed of Appendix~\ref{app:full-sweep}), positive when ASMI is ahead, with asterisks marking intervals that exclude zero. The first two $\Delta$ columns follow the column's top ASMI variant and the last reports the sampling-free Sem-ASMI against Semantic Entropy (SemEnt), the pre-specified primary reference. The per-column best baseline is the conservative reference, taken over all 17 baselines, and RAUQ-E denotes the entropy variant of RAUQ. Cluster counts are $499$ on CoQA, $1{,}934$ on SQuAD, $999$ on BabiQA, and $4{,}404$ on TriviaQA. Point PRR values agree with the full benchmark tables within their standard errors.}
\label{tab:h2h}
\end{table*}

\paragraph{Instability of depth selection}
BabiQA is the one grounded benchmark whose PRR is non-monotonic in depth, with a pronounced dip at $80\%$ on the two Qwen backbones. For unweighted ASMI on Qwen3-8B it drops from $0.78$ at $70\%$ to $0.48$ at $80\%$ and recovers at $90\%$, instead of following the smooth CoQA trend. The task explains this. BabiQA answers are a single location word, so the sequence score is carried by essentially one token and there is almost no averaging over positions to smooth depth effects, unlike CoQA where up to twenty tokens are averaged. The task is also solved by a localized copy of the last-mentioned location, so whether a given layer is masked interacts sharply with whether that routing step is perturbed, which produces a bad layer near $80\%$ for the Qwen models. The small evaluation set ($n = 1\mathrm{K}$) then widens the standard error and amplifies these swings. Semantic weighting cushions the dip, raising the Qwen3-8B $80\%$ score from $0.48$ for ASMI to $0.70$ for Sem-ASMI, which indicates that much of the disagreement introduced at that layer is surface-form and is discounted by the kernel.
 
Llama-2-7B is low at every depth on BabiQA for a distinct reason that the mechanism analysis pins down. Its generations are well-formed single location words, but head masking barely perturbs its output on this task. The token-level $\mathrm{MI}_t$ has median $0.000$ and mean $0.002$, against $0.025$ on Qwen3-8B, so the model is over-robust to the perturbation and insensitive along the attention path. Correct and incorrect answers therefore carry the same near-zero MI and their score distributions overlap almost completely (Fig.~\ref{fig:error-separation}, top). ASMI retains only an attenuated ranking, while the output-distribution MSP still separates them (Table~\ref{tab:babiqa-mechanism}). On Qwen3-8B, where the model routes the task and wrong answers are more fragile, the wrong-answer distribution shifts upward (Fig.~\ref{fig:error-separation}, bottom) and ASMI matches MSP. The MI magnitude flags this over-robustness before any correctness label is available. At the representative $60\%$ depth BabiQA is well-behaved on three of four backbones and enters the main comparison, but its depth profile is the least stable of the grounded family, consistent with the underpowered head-criticality coupling in the mechanism analysis.

\section{Head-to-Head Significance}
\label{app:h2h}
The bootstrap standard errors of the main tables overstate the uncertainty of a comparison, because every method scores the same greedy responses under the same correctness labels. We therefore test differences directly with a cluster-respecting paired bootstrap. Each of the $B = 10\mathrm{K}$ resamples draws one set of clusters with replacement, using the same cluster keys and seed as Appendix~\ref{app:full-sweep}, and applies the same indices to both methods, so the shared per-example variation cancels and the intervals are narrower than the marginal standard errors imply, by roughly half on CoQA. We report $\Delta$PRR with a 95\% percentile interval and call a comparison a tie when the interval includes zero. Semantic Entropy is the pre-specified primary reference, fixed before running the test to avoid selecting the winner after the fact, and the per-column best baseline is the conservative reference, since the maximum over all 17 baselines is the hardest comparison for ASMI.
Table~\ref{tab:h2h} reports all sixteen columns. Against Semantic Entropy the top ASMI variant wins significantly in six of the twelve grounded columns, loses in two, and ties in four. Against the per-column best baseline ASMI leads in eight grounded columns, significantly in three (CoQA on both Qwen backbones and BabiQA on Qwen3-8B), and the remaining leads are ties. A tie carries information of its own here, since in every tied column the sampling-free Sem-ASMI ties or beats Semantic Entropy from the greedy response alone (last column of Table~\ref{tab:h2h}). One of the two losses against Semantic Entropy is the over-robust Llama-2-7B BabiQA cell the operating envelope already maps, and the other is Qwen3-8B on SQuAD. The largest single gap in the table, the $+0.236$ of the Llama-2-7B SQuAD column, coincides with the column where $28.5\%$ of greedy generations are empty (Appendix~\ref{app:full-benchmark}), so we checked the tallies without it: over the remaining eleven grounded columns the top variant still wins five, loses two, and ties four against Semantic Entropy, and still leads the per-column best baseline in eight with three significant, so no headline count depends on this column. On parametric TriviaQA ASMI sits significantly below the best baseline in three of the four columns, so the designed boundary holds under the paired test as well.
\section{Estimator Stability}
\label{app:stability}
\paragraph{Protocol.}
The bootstrap SEs of the main tables measure sampling variability over test examples and treat each method's scores as fixed. They do not measure the variability of the estimator itself, whose scores depend on internal randomness, namely sampled head masks for ASMI and stochastically decoded samples for sample-diversity baselines. We fix a 500-example CoQA subset, the backbone, and the scored greedy responses, and redraw only the method-internal randomness five times, using disjoint sets of $S = 40$ masks for the ASMI variants and independent resamplings of the $10$ stochastic generations (temperature $0.5$ following~\citealp{duan2024shifting}, as configured in all main experiments) for the baselines. Because the examples are fixed, differences across redraws reflect measurement noise alone. Absolute PRR levels differ from the main tables (subset, single layer). This experiment measures variance, not level.
\begin{table}[ht!]
\centering
\footnotesize
\setlength{\tabcolsep}{3.5pt}
\begin{tabular}{lcccc}
\toprule
Method & mean & sd & range & rank-$\rho$ \\
\midrule
Adapt-ASMI & $.493$ & $.006$ & $[.486, .501]$ & $.98$ \\
Sem-ASMI   & $.491$ & $.004$ & $[.486, .496]$ & $.98$ \\
ASMI       & $.444$ & $.011$ & $[.432, .461]$ & $.96$ \\
\midrule
Semantic Entropy        & $.474$ & $.020$ & $[.455, .505]$ & $.94$ \\
SAR                     & $.477$ & $.014$ & $[.459, .493]$ & $.92$ \\
MC Sequence Entropy     & $.450$ & $.014$ & $[.440, .470]$ & $.95$ \\
Kernel Language Entropy & $.427$ & $.041$ & $[.373, .479]$ & $.81$ \\
\bottomrule
\end{tabular}
\caption{Estimator stability: PRR across five redraws of method-internal randomness on a fixed 500-example CoQA subset. rank-$\rho$ is the mean pairwise Spearman correlation of per-example scores across redraws.}
\label{tab:stability}
\end{table}
\paragraph{Results.}
Table~\ref{tab:stability} reports the across-redraw standard deviation of PRR and the mean pairwise Spearman correlation of the per-example scores. Adapt-ASMI varies by $\pm 0.006$ across redraws and Sem-ASMI by $\pm 0.004$, two to five times less than the sample-diversity baselines (from $\pm 0.014$ to $\pm 0.041$), and their example rankings are nearly deterministic (rank-$\rho = 0.98$). This follows from the design: ASMI scores a \emph{fixed} greedy response under structural perturbations whose head coverage is guaranteed (Appendix~\ref{app:coverage-fidelity}), whereas sample-diversity scores inherit the stochasticity of temperature decoding. With five redraws the sd estimates are coarse. Ranges are reported alongside.
\section{Benchmark Prompts and Examples}
\label{app:examples}
For reproducibility we show the prompt format and one representative example per benchmark, exactly as produced by the LM-Polygraph pipeline on base models with no chat template. Long contexts are truncated with [...], and the gold answer follows each block.
\paragraph{CoQA (grounded, 0-shot).}
A flattened story followed by a single question, with the answer stated in the story.
\begin{lstlisting}[numbers=none,basicstyle=\scriptsize\ttfamily]
The following are stories and questions about them. Each story is
followed by a question and answer to a given question.
Story: Once upon a time, in a barn near a farm house, there lived a
little white kitten named Cotton. [...] But she was the only white
one in the bunch. [...] Then Cotton thought, "I like being special".
Question: What color was Cotton?
Answer:
\end{lstlisting}
Gold answer: \texttt{white}.
\paragraph{SQuAD (grounded, 0-shot).}
A passage with an extractive question whose answer is a span of the passage.
\begin{lstlisting}[numbers=none,basicstyle=\scriptsize\ttfamily]
Read the following passage and answer the question.
Passage:
Super Bowl 50 was an American football game to determine the champion
of the National Football League (NFL) for the 2015 season. The AFC
champion Denver Broncos defeated the NFC champion Carolina Panthers
24-10 to earn their third Super Bowl title. [...]
Question: Which NFL team represented the AFC at Super Bowl 50?
Answer:
\end{lstlisting}
Gold answer: \texttt{Denver Broncos}.
\paragraph{BabiQA (grounded, 1-shot).}
A synthetic location-tracking task with a fixed one-shot exemplar. The answer is a single word copied from the context.
\begin{lstlisting}[numbers=none,basicstyle=\scriptsize\ttfamily]
Imagine that you are only able to say a single word. Answer the
question given a context. You must only output the full name of the
location the same way it is mentioned in the text.
Example:
Context: Mary moved to the bathroom. John went to the hallway. [...]
Question: Where is Sandra?
Answer: bathroom
Context: John travelled to the hallway. Mary journeyed to the bathroom.
Question: Where is John?
Answer:
\end{lstlisting}
Gold answer: \texttt{hallway}.
\paragraph{TriviaQA (parametric, closed-book, 5-shot).}
Five fixed exemplars followed by the query, with no passage provided.
\begin{lstlisting}[numbers=none,basicstyle=\scriptsize\ttfamily]
Question: What is the capital of the Indian state of Tamil Nadu?
Answer:chennai
Question: Who rules Narnia following the reign of High King Peter [...]?
Answer:prince caspian
[... three more fixed exemplars ...]
Question: Who was the man behind The Chipmunks?
Answer:
\end{lstlisting}
Gold answer: \texttt{David Seville}, which must come from parametric knowledge.
\end{document}